\RequirePackage{fix-cm}
\documentclass[twocolumn]{svjour3}          
\smartqed  
\usepackage{algorithm}
\usepackage{algorithmic}
\usepackage{graphicx}
\usepackage{amssymb,natbib,amsmath}
\usepackage[colorlinks,citecolor=red,urlcolor=red, draft]{hyperref}
\DeclareGraphicsExtensions{.pdf,.jpeg,.png,.jpg}
\usepackage[caption=false,font=footnotesize]{subfig}

\usepackage{breakcites}
\usepackage{etoolbox}
\makeatletter
\patchcmd{\@citex}{,}{;}{}{}
\makeatother

\begin{document}

\title{Long-Term Online Multi-Session Graph-Based SPLAM with Memory Management\thanks{This work was supported by the Natural Sciences and Engineering Research Council of Canada (NSERC), the Canada Research Chair program and the Canadian Foundation for Innovation.}
}


\author{Mathieu Labb\'e         \and
        Fran{\c c}ois Michaud 
}


\institute{M. Labb\'e \at
              \email{mathieu.m.labbe@usherbrooke.ca}             
              \and
              F. Michaud \at
              \email{francois.michaud@usherbrooke.ca} 
\and
  Interdisciplinary Institute for Technological Innovation (3IT), Universit\'{e} de Sherbrooke, Sherbrooke, Qu\'{e}bec, Canada
}

\date{}

\maketitle

\sloppy 

\begin{abstract}
For long-term simultaneous planning, localization and mapping (SPLAM), a robot should be able to continuously update its map according to the dynamic changes of the environment and the new areas explored. 
With limited onboard computation capabilities, a robot should also be able to limit the size of the map used for online localization and mapping. 
This paper addresses these challenges using a memory management mechanism, which identifies locations that should remain in a Working Memory (WM) for online processing from locations that should be transferred to a Long-Term Memory (LTM). 
When revisiting previously mapped areas that are in LTM, the mechanism can retrieve these locations and place them back in WM for online SPLAM.  
The approach is tested on a robot equipped with a short-range laser rangefinder and a RGB-D camera, patrolling autonomously 10.5 km in an indoor environment over 11 sessions while having encountered 139 people.

\end{abstract}

\section{Introduction}

The ability to simultaneously map an environment, localize itself in it, and plan paths using this information is known as \textit{Simultaneous Planning, Localization And Mapping}, or SPLAM \citep{stachniss2009robotic}.
This task can be particularly complex when done online on a robot with limited computing resources in large, unstructured and dynamic environments. Since SPLAM can be seen as an extension of \textit{Simultaneous Localization And Mapping} (SLAM), many approaches exist \citep{thrun05}. 
Our interest lies with graph-based SLAM approaches \citep{grisetti2010tutorial}, for which combining a lightweight topological map over a detailed metrical map reveals to be more suitable for large-scale mapping and navigation \citep{konolige2011navigation}. 

Two important challenges in graph-based SPLAM are :

\begin{itemize}
\item Multi-session mapping, also known as the \textit{kidnapped robot problem} or the \textit{initial state problem}: when  turned on, a robot does not know its relative position to a map previously created, making it impossible to plan a path to a previously visited location.
A solution is to have the robot localize itself in a previously-built map before initiating mapping. 
This solution has the advantage of always using the same referential, resulting in only one map is created across the sessions. 
However, the robot must start in a portion already mapped of the environment. 
Another approach is to initialize a new map with its own referential on startup, and when a previously visited location is encountered, a transformation between the two maps can be computed. 
The transformations between the maps can be saved explicitly with special nodes called \textit{anchor nodes} \citep{mcdonald2012real, kim2010multiple}, or implicitly with links added between each map \citep{konolige2009towards, latif2013robust}. 
This process is referred to as \textit{loop closure detection}.
Loop closure detection approaches that are independent of the robot's estimated position \citep{Ho06} can intrinsically detect if the current location is a new location or a previously visited one among all the mapping sessions conducted in the past. 
Popular loop closure detection approaches are appearance-based \citep{GarciaFidalgo20151}, exploiting the distinctiveness of images of the environment. 
The underlying idea is that loop closure detection is done by comparing all previous images with the new one. 
When loop closures are found between the maps, a global map can be created by combining the maps from each session. 
In graph-based SLAM, graph pose optimization approaches \citep{Folkesson07, Grisetti07a, kummerle11g2o, johannsson2013temporally} use these loop closures to reduce odometry errors inside each map and in between the maps. 

\item Long-term mapping in dynamic environments. \textit{Persistent} \citep{milford2010persistent}, \textit{lifelong} \citep{konolige2009towards} or \textit{continuous} \citep{pirker11cdslam} 
are terms generally used to describe SLAM approaches working in such conditions. Continuously updating and adding new data to the map in unbounded or dynamic environments will inevitably increase the map size over time. 
Online simultaneous planning, localization and mapping requires that new incoming data be processed faster than the time to acquire them. 
For example, if data are acquired at 1 Hz, updating the map should be done in less than 1 sec. 
As the map grows, the time required for loop closure detection and graph optimization increases, and eventually limits the size of the environment that can be mapped and used online. 
\end{itemize}

To address these challenges, we introduce SPLAM-MM, a graph-based SPLAM with a memory management (MM) mechanism. 
As demonstrated in \citep{labbe13appearance}, memory management can be used to limit the size of the map so that loop closure detections are always processed under a fixed time limit, thus satisfying online requirements for long-term and large-scale environment mapping. 
The idea behind SPLAM-MM is to limit the number of nodes available for loop closure detection and graph optimization, keeping enough observations in the map for successful online localization and planning while still having the ability to generate a global representation of the environment that can adapt to changes over time. 

The paper is organized as follows. 
Section \ref{sec:related_work} reviews graph-based SLAM approaches that reduce the size of the map when revisiting the same environment while continuously adapting to dynamic changes.
Section \ref{sec:principles} describes the implementation and the operating principles associated with the use of memory management with a graph-based SPLAM approach, which extends our previous metric-based SLAM approach \citep{labbe14online} with a new planning capability.  
The implementation integrates four algorithms: loop closure detection \citep{labbe13appearance}, graph optimization \citep{Grisetti07a}, metrical path planner \citep{officemarathon} and a custom topological path planner. 
Section \ref{sec:results} presents experimental results of 11 SPLAM sessions using the AZIMUT-3 robot in an indoor environment over 10.5 km. 
Section \ref{sec:discussion} discusses strengths and limitations of SPLAM-MM, and Section \ref{sec:conclusion} concludes the paper.

\section{Related Work}
\label{sec:related_work}

Lifelong appearance-based SLAM requires dealing with dynamic environments.
\cite{glover10} present an appearance-based SLAM approach that had to operate in different lighting conditions over three weeks. 
An interesting observation from their experiments is that even when revisiting the same locations, the map still grows: in dynamic environments, the loop closure detector is sometimes unable to detect loop closures, duplicating locations in the map. 
A map management approach is therefore required to limit map size.
In highly dynamic environments, multiple views of the same location may also be required for proper localization. \cite{churchill2012practice} present a graph-based SLAM approach where visual experiences of the same locations are kept in the map, to increase localization robustness to dynamic changes caused for instance by outdoor illumination conditions. 
If localization fails when revisiting an area, new experiences are added to the map. 
Even if adding new visual experiences to the map happens less often over time (as the robot explores the same location), there is no mechanism to limit this.
\cite{pirker11cdslam} present a continuous monocular SLAM approach where new key frames are added to the map only when the environment has changed, to keep its size proportional to the explored space. 
But if the environment changes very often, there is no mechanism to limit the number of key frames over the same physical location. 

Some SLAM approaches can handle dynamic changes of the environment while limiting the size of the map for long-term operation. 
\cite{biber2005dynamic} present a sample-based representation for maps, to handle changes at different timescales, tracking both stationary and non-stationary elements of the environment. 
The idea is to refresh samples stored for each timescale with new sensor measurements. 
Map growth is then indirectly limited as older memories fade at different rates depending on the timescale. 
\cite{walcott2012dynamic} describe Dynamic Pose-Graph SLAM (DPG-SLAM), a long-term mapping approach that detects static and dynamic changes of the environment through time. 
To keep consistency of the graph while reducing its size, nodes that are not observable anymore are removed. 
\cite{johannsson2013temporally} also remove unobservable nodes to limit the size of the map over time when revisiting the same area. 
Similar nodes of the graph are merged together while keeping only the new loop closure detection. However, the graph size is not bounded when exploring new areas. 
\cite{krajnik2016persistent} present an occupancy grid approach where each cell in the map estimates its occupancy value depending on periodical and cyclic changes occurring in the environment. This increases localization and navigation accuracy in dynamic environments compared to static maps, as the predicted map represents the correct state of the environment at that time of the day (e.g., doors can change to be opened or closed). The maximum data kept for each cell is bounded by some parameters (depending on the smallest and longest cyclic periods that should be detected), thus keeping memory usage fixed. However, the approach assumes that the navigation phase always occur in the same environment as the first mapping cycle, without possibility to extend it afterward.

These problems of lifelong SLAM are also addressed in some SPLAM approaches. 
\cite{milford2010persistent} present a solution to limit the size of the map (called experience map) while revisiting the same area: close nodes are merged together up to a maximum density threshold. 
This approach has the advantage of making the map size independent of the operating time, but the diversity of the observations on each location is somewhat lost. 
\cite{konolige2011navigation} use a view-based graph SLAM approach \citep{konolige2009towards} in a SPLAM context. 
The approach preserves diversity of the images referring to the same location so that the map can handle dynamic changes over time, and forgetting images limits the size of the graph over time when revisiting the same area. 
However, the graph still grows when visiting new areas. 

Overall, these approaches reduce map size when revisiting the same area, while continuously adapting to dynamic changes. 
This makes them independent or almost independent of the operation time of the robot in these conditions, but they are all limited to a maximum size of the environment that can be mapped online. 
The SPLAM-MM approach deals specifically with this limitation.

\section{Memory Management for SPLAM}
\label{sec:principles}
\begin{figure}[!t] 
\centering 
\includegraphics[width=\columnwidth]{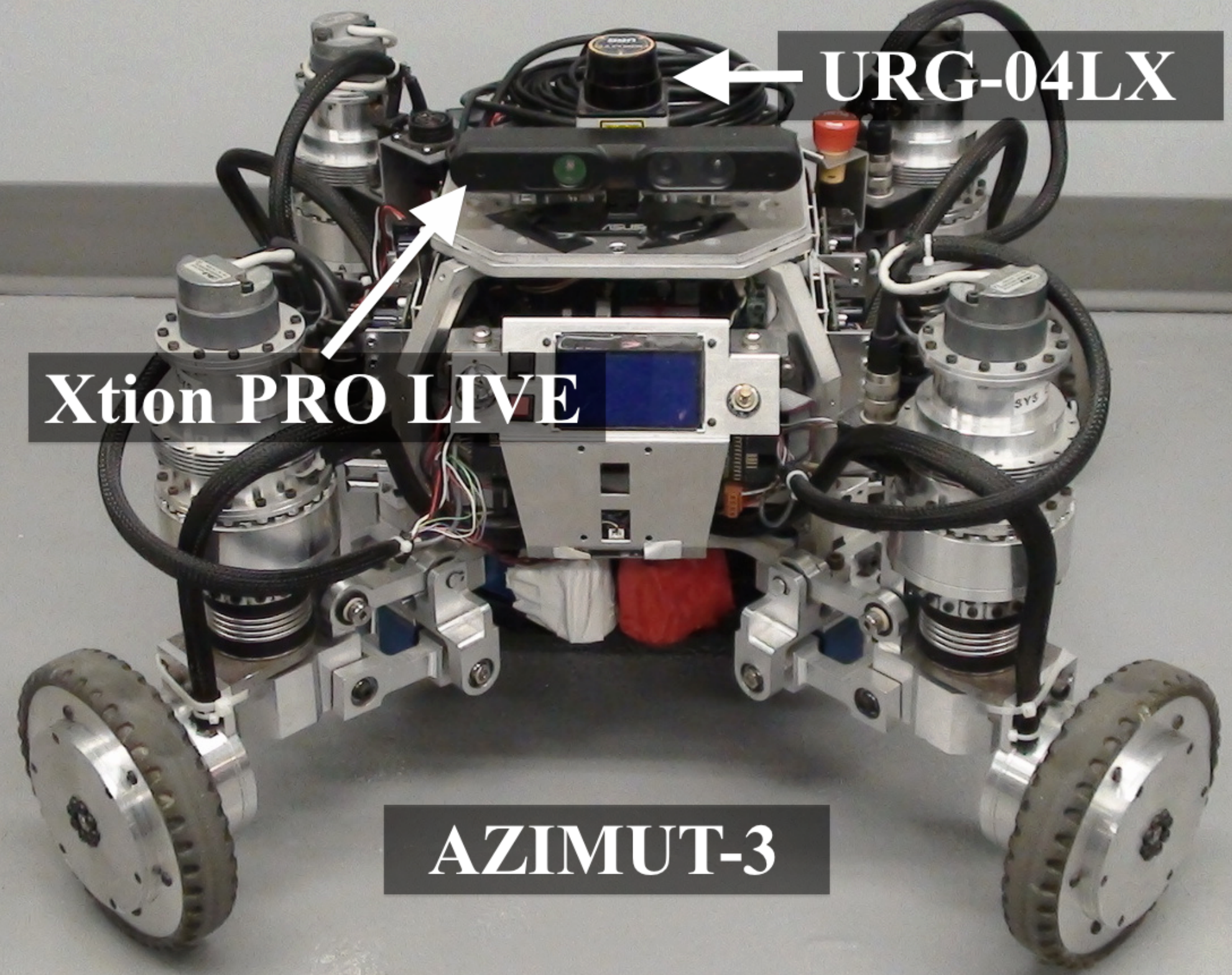} 
\caption{The AZIMUT-3 robot equipped with a URG-04LX laser range finder and a Xtion PRO LIVE sensor.} 
\label{fig:azimut} 
\end{figure}

\begin{figure*}[!t] 
\centering 
\includegraphics[width=\textwidth]{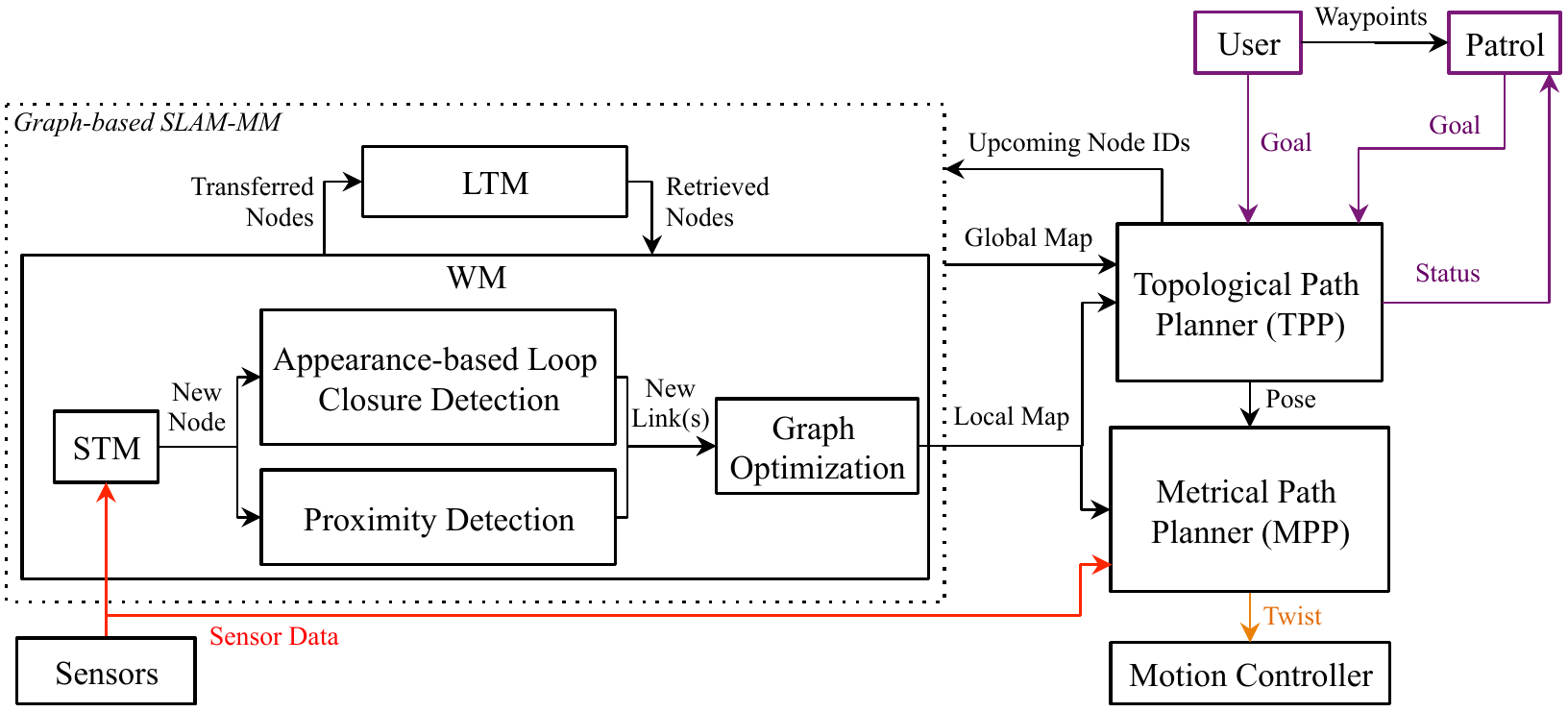} 
\caption{Memory management and control architecture of SPLAM-MM.} 
\label{fig:architecture} 
\end{figure*}

The underlying representation of SPLAM-MM  is a graph with nodes and links. 
The nodes contain the following information: 

\begin{itemize}

\item ID: unique time index of the node.
\item Weight: an indication of the importance of the node, used for memory management.
\item Bag-of-words (BOW): visual words used for loop closure detections. 
They are SURF features \citep{Bay08} quantized to an incremental vocabulary based on KD-Trees.
\item Sensor data: used to find similarities between nodes and to construct maps. For this paper, our implementation of SPLAM-MM is using the AZIMUT-3 robot \citep{ferland10teleopration}, equipped with an URG-04LX laser rangefinder and a Xtion Pro Live RGB-D camera, as shown by Fig. \ref{fig:azimut}. The sensory data used are: 

\begin{itemize}
\item Pose: the position of the robot computed by its odometry system (e.g., the value given by wheel odometry), expressed in  ($x, y, \theta$) coordinates. 
\item RGB image: used to extract visual words.
\item Depth image: used to find 3D position of the visual words. The depth image is registered with the RGB image, i.e., each depth pixel corresponds exactly to the same RGB pixel.
\item Laser scan: used for loop closure transformations and odometry refinements, and by the Proximity Detection module. 
\end{itemize}

\end{itemize}

The links store rigid transformations (i.e., Eucledian transformation derived from odometry or loop closures) between nodes. 
There are four types of links: 

\begin{itemize}
\item Neighbor link: created between a new node and the previous one. 
\item Loop closure link: added when a loop closure is detected between the new node and one in the map. 
\item Proximity link: added when two close nodes are aligned together. 
\item Temporary link: used for path planning purposes. It is used to keep the planned path connected to the current map.
\end{itemize}

Figure \ref{fig:architecture} presents a high-level representation of SPLAM-MM. 
Basically, it consists of a graph-based SLAM module with memory management, to which path planners are added. 
Memory management involves the use of a Working Memory (WM) and a Long-Term Memory (LTM).
WM is where maps, which are graphs of nodes and links, are processed. 
To satisfy online constraints, nodes can be transferred and retrieved from LTM.
More specifically, the WM size indirectly depends on a fixed time limit $T$: when the time required to update the map (i.e., the time required to execute the processes in the Graph-based SLAM-MM block) reaches $T$, some nodes of the map are transferred from WM to LTM, thus keeping WM size nearly constant and processing time around $T$. 
However, when a loop closure is detected, neighbors in LTM with the loop closure node can be retrieved from LTM to WM for further loop closure detections. 
In other words, when a robot revisits an area which was previously transferred to LTM, it can incrementally retrieve the area if a least one node of this area is still in WM. 
When some LTM nodes are retrieved, nodes in WM from other areas in the map can be transferred to LTM, to limit map size in WM and therefore keeping processing time around $T$.

Therefore, the choice of which nodes to keep in WM is key in SPLAM-MM. 
The objective is to have enough nodes in WM from each mapping session for loop closure detections and to keep a maximum number of nodes in WM for generating a map usable to follow correctly a planned path, while still satisfying online processing.
Two heuristics are used to establish the compromise between selection of which nodes to keep in WM and online processing: 

\begin{itemize}
\item Heuristic 1 is inspired from observations made by psychologists \citep{atkinson1968human, baddeley1997human} that people remember more the areas where they spent most of their time, compared to those where they spent less time. 
In terms of memory management, this means that the longer the robot is at a particular location, the larger the weight of the corresponding node should be. 
Oldest and less weighted nodes in WM are transferred to LTM before the others, thus keeping in WM only the nodes seen for longer periods of time. 
As demonstrated in \citep{labbe13appearance}, this heuristic reveals to be quite efficient in establishing the compromise between search time and space, as driven by the environment and the experiences of the robot. 

\item Heuristic 2 is used to identifies nodes that should stay in WM for autonomous navigation. 
Nodes on a planned path could have small weights and may be identified for transfer to LTM by Heuristic 1, thus eliminating the possibility of finding a loop closure link or a proximity link with these nodes and correctly follow the path. 
Therefore, Heuristic 2 must supersede Heuristic 1 and allow upcoming nodes to remain in WM, even if they are old and have a small weight. 

\end{itemize}

The Graph-based SLAM-MM block provides two types of maps derived from nodes in WM and LTM: 

\begin{itemize}
\item Local map, i.e., the largest connected graph that can be created from the last node in WM with nodes available in WM only. The local map is used for online path planning.
\item Global map, i.e., the largest connected graph that can be created from the last node in WM with nodes in WM and LTM. 
It is used for offline path planning.
\end{itemize}

\begin{figure}[!t] 
\centering 
\includegraphics[width= \columnwidth]{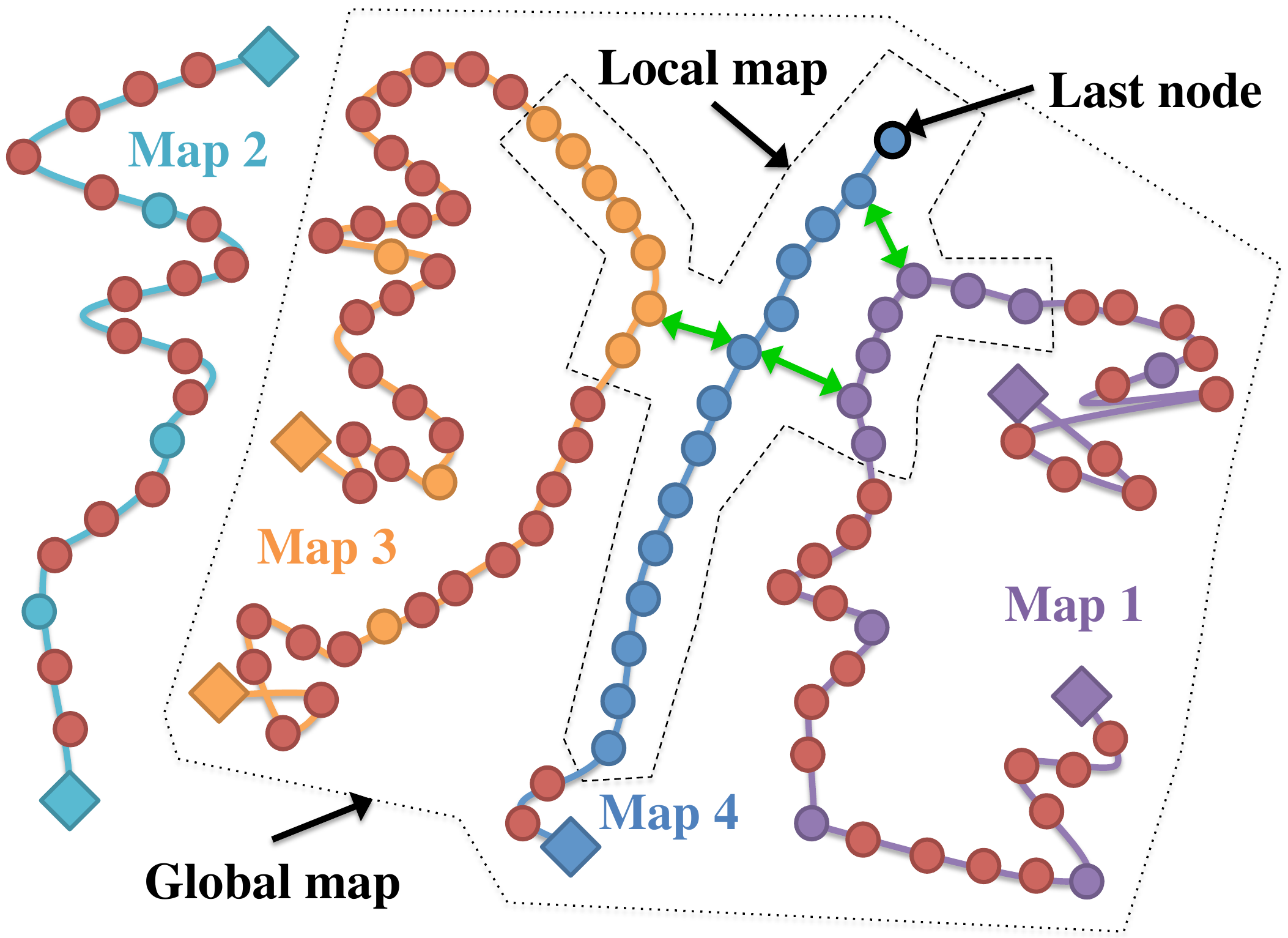} 
\caption{Illustration of the local map (inner dashed area) and the global map (outer dotter area) in multi-session mapping. Red nodes are in LTM, while all other nodes are in WM. Loop closure links are shown using bidirectional green arrows.} 
\label{fig:tree} 
\end{figure}

Figure \ref{fig:tree} uses diamonds to represent initial and end nodes for each mapping session. 
The nodes in LTM are shown in red and the others are those in WM. 
The local map is created using only the nodes in WM that are linked to the last node. 
The graph linking the last node with other nodes in WM and LTM represents the global map (outer dotted area). 
If loop closure detections are found between nodes of different maps, loop closure links can be generated, and the local map can span over multiple mapping sessions. 
Other nodes in WM but not included in the local map are unreachable from the last node, but they are still used for loop closure detections since all nodes in WM (including those in Map 2 for instance) are examined. 

The modules presented in Fig. \ref{fig:architecture} are described as follows. 

\subsection{Short-Term Memory Module}
\label{STM}
Short-Term Memory (STM) is the entry point where sensor data are assembled into a node to be added to the map. Similarly to \citep{labbe13appearance}, the role of the STM module is to update node weight based on visual similarity. 
When a node is created, a unique time index ID is assigned and its weight is initialized to 0. 
The current pose, RBG image, depth image and laser scan readings are also memorized in the node.
If two consecutive nodes have similar images, i.e., the ratio of corresponding visual words between the nodes is over a specified threshold $Y$, the weight of the previous node is increased by one. 
If the robot is not moving (i.e., odometry poses are the same), the new node is deleted. 
To reduce odometry errors on successive STM nodes, transformation refinement is done using 2D iterative-closest-point (ICP) optimization \citep{besl1992method} on the rigid transformation of the neighbor link with the previous node and the corresponding laser scans. 
If the ratio of ICP point correspondences between the laser scans over the total laser scan size is greater or equal to $C$, the neighbor link's transformation is updated with the correction. 

When the STM size reaches a fixed size limit of $S$ nodes, the oldest node in STM is moved to WM. 
STM size is determined based on the velocity of the robot and at which rate the nodes are added to the map. 
Images are generally very similar to the newly added node, keeping $S$ nodes in STM avoids using them for appearance-based loop closure detection once in WM. 
For example, at the same velocity, STM size should be larger if the rate at which the nodes are added to map increases, in order to keep nodes with consecutive similar images in STM. 
Transferring nodes with images very similar with the current node from STM to WM too early limits the ability to detect loop closures with older nodes in WM.

\subsection{Appearance-based Loop Closure Detection Module}
\label{loopclosuredetection}
Appearance-based loop closure detection is based on the bag-of-words approach described in \citep{labbe13appearance}. 
Briefly, this approach uses a bayesian filter to evaluate appearance-based loop closure hypotheses over all previous images in WM. 
When a loop closure hypothesis reaches a pre-defined threshold $H$, a loop closure is detected. 
Visual words of the nodes are used to compute the likelihood required by the filter. In this work, the Term Frequency-Inverse Document Frequency (TF-IDF) approach \citep{sivic2003video} is used for fast likelihood estimation, and FLANN (Fast Library for Approximate Nearest Neighbors) incremental KD-Trees \citep{muja_flann_2009} are used to avoid rebuilding the vocabulary at each iteration. 
To keep it balanced, the vocabulary is rebuilt only when it doubles in size.

The RGB image, from which the visual words are extracted, is registered with a depth image. 
Using (\ref{eq:depth}), for each 2D point $(x,y)$ in the rectified RGB image, a 3D position $P_{xyz}$ can be computed using the calibration matrix (focal lengths $f_x$ and $f_y$, optical centres $c_x$ and $c_y$) and the depth information $d$ for the corresponding pixel in the depth image. 
The 3D positions of the visual words are then known. 
When a loop closure is detected, the rigid transformation between the matching images is computed using a RANSAC (RANdom SAmple Consensus) approach which exploits the 3D visual word correspondences \citep{Rusu_ICRA2011_PCL}. 
If a minimum of $I$ inliers are found, the transformation is refined using the laser scans in the same way as the odometry correction in STM using 2D ICP transformation refinement. 
If transformation refinement is accepted, then a loop closure link is added with the computed transformation between the corresponding nodes. 
The weight of the current node is updated by adding the weight of the loop closure hypothesis node and the latter is reset to 0, so that only one node with a large weight represents the same location. 

\begin{equation}
\label{eq:depth}
P_{xyz} =  \left[\frac{(x-c_x)\cdot d}{f_x}, \frac{(y-c_y)\cdot d}{f_y}, d\right]^T
\end{equation}

By doing appearance-based loop closure detection this way, setting $H$ high means that there is less chance of detecting false positives, but at the cost of detecting less loop closures \citep{labbe13appearance}. 
For SPLAM-MM, $H$ can be set relatively low to detect more loop closures because false positives that are geometrically different will be rejected by the rigid transformation computation step (i.e., the 3D visual word correspondences and 2D ICP transformation refinement). 

\subsection{Proximity Detection Module}  
\label{localloopclosuredetection}
\begin{figure*}[!t] 
\centering 
\includegraphics[width= 6in]{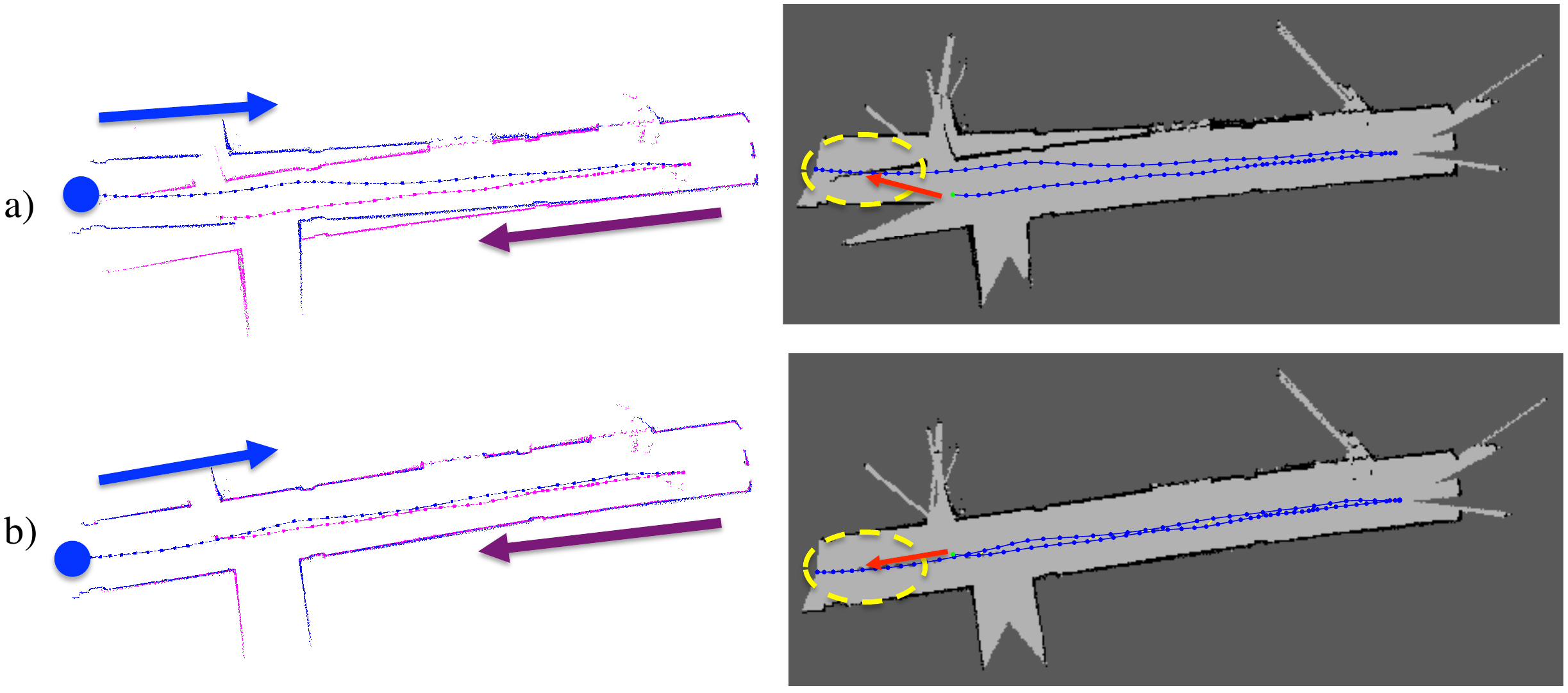} 
\caption{Illustration of the role of the Proximity Detection module. On the left are the raw laser scans, the blue dot is the starting position, and on the right the corresponding occupancy grid map at $0.05$ m resolution (black, light gray and dark gray areas are occupied, empty and unknown spaces, respectively). In a), the yellow circle on the right locates the problematic situation: after the second traversal, the first nodes of the graph are located exactly over the wall, making it impossible to plan a path (red arrow on the right) to return to the starting position. In b), proximity links are detected using only the laser scans, and the local map can then be correctly optimized. } 
\label{fig:hallproblem} 
\end{figure*}

Appearance-based loop closure detection is limited by the perceptual range of the sensory data used.
For instance, when the robot is revisiting areas in opposite direction, the RGB-D camera on AZIMUT-3 is not pointing in the same direction compared to when the nodes were created, and thus no appearance-based loop closures can be detected. 
This also happens when there are not enough visual features under the depth range of the RGB-D camera (e.g., white walls or long halls). 
Simply relying on appearance-based loop closure detections for map corrections would then limit path planning capabilities, and make navigation difficult in such conditions.
Figure \ref{fig:hallproblem}a illustrates a situation where the robot is in a hall coming back to its starting position in reverse direction. 
Setting a goal at the starting position would make the planner fail because no loop closures could be found to correct the odometry, resulting in having a wall directly placed on the starting position. 
One solution would be to have the robot visit the nodes of the graph backward so loop closures could be detected to correct the map, and ultimately be able to reach the starting position. 
However, it is inefficient and unsafe if the robot does not have sensors pointing backward. 
To deal with such situations, the Proximity Detection module uses laser rangefinder data to correct odometry drift in areas where the camera cannot detect loop closures. 
With a field of view of more than $180^\circ$, the laser scans can be aligned in reverse direction, generating proximity links. 
As laser scans are not as discriminative as images, proximity detection is restricted to nodes of the local map located around the estimated position of the robot. 
Figure \ref{fig:hallproblem}b illustrates the result.

\begin{figure}[!t] 
\centering 
\includegraphics[width=\columnwidth]{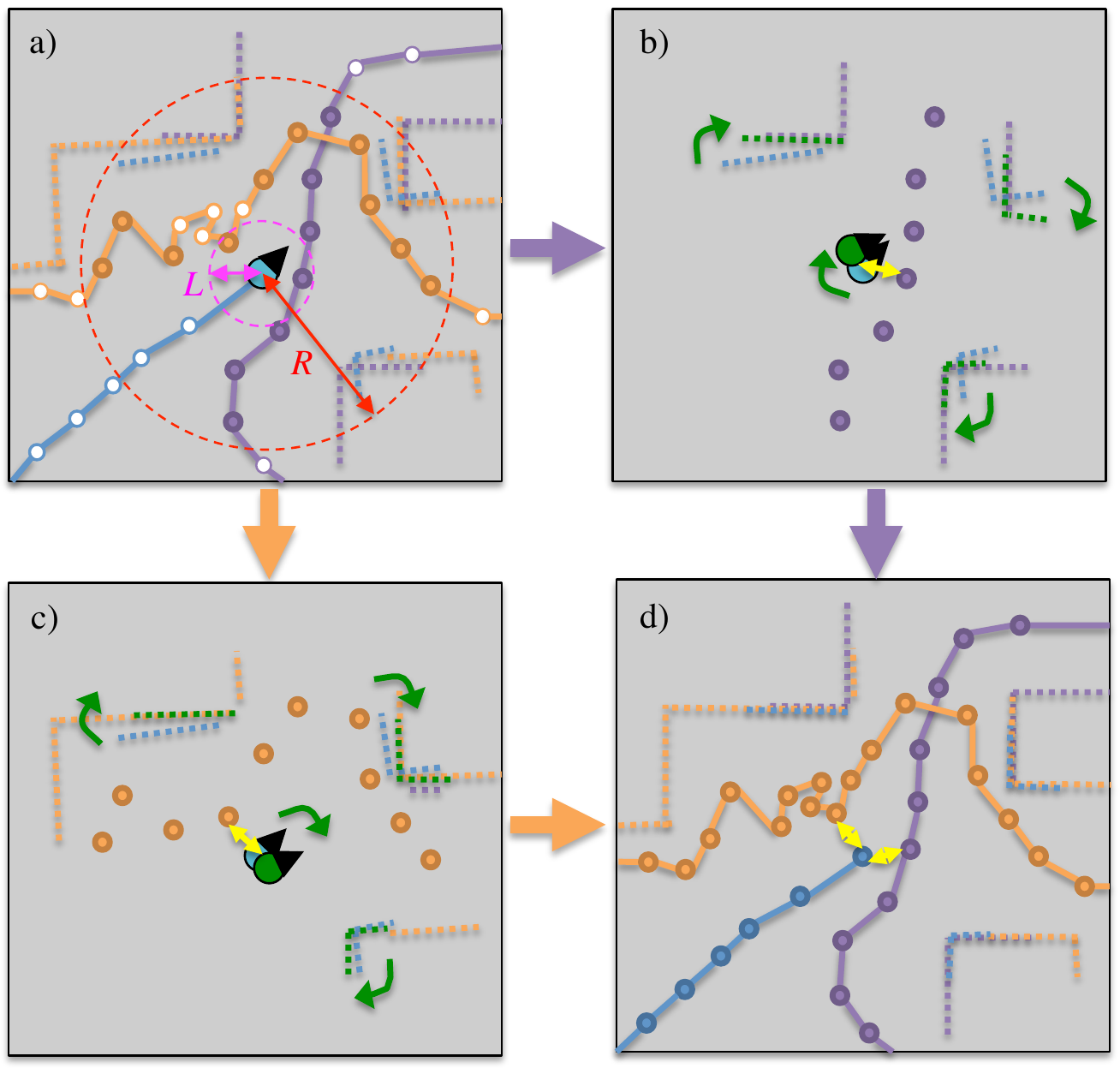} 
\caption{Illustration of how proximity detection works. In a), the larger dashed circle represents the radius $R$ used to determine close-by nodes, and the smaller dashed circle defined by $L$ is used to limit the length of the links to be created. 
The empty dots are nodes for which the laser scans are not used, either because they are outside the radius $R$, they are too close from each other or they are in STM. In b) and c), nodes in the radius $R$ from the two segmented groups of nodes are processed for proximity detection. In d), proximity links are added (yellow), and after graph optimization, the groups of nodes are connected together and the respective laser scans are now aligned.
} 
\label{fig:scanMatching} 
\end{figure}

Figure \ref{fig:scanMatching} illustrates how nodes located close to the robot are selected by the Proximity Detection module. 
Only nodes in the local map with their pose inside radius $R$ centered on the robot are used. 
Nodes in STM are not considered in order to avoid adding useless links with nodes close by: this would increase graph optimization time without adding significative improvements of the map. 
The nodes are then segmented into groups with nodes connected only by neighbor links. 
A group must have its nearest node from the robot inside a fixed radius $L$ defining close-by nodes (with $L<R$) to be considered for proximity detection, to keep the length of the resulting proximity links small for path planning. 
Note that Appearance-based Loop Closure Detection is done before Proximity Detection, thus if the nearest node has already a loop closure with the new node, the group is ignored. 
Proximity detection is then applied separately on each group of nodes by doing the following steps:

\begin{enumerate} 
\item A rigid transformation between the nearest node of each group and the new node added to map is computed as in Section \ref{loopclosuredetection}, and if it is accepted, a proximity link is added between the corresponding nodes, and the group of nodes is ignored for step 2. These links are referred as visual proximity links because visual words are used in the transformation estimation.
\item To avoid having to compare multiple nodes with very similar laser scans (and thus to save computation), only the more recent node among those in the same fixed small radius $L$ (centered on each node) is kept along the nodes in a remaining group.
Then for each group, the laser scans of the nodes are merged together using their respective pose. 
2D ICP transformation refinement is done between the merged laser scans and the one of the new node. 
If the transformation is accepted, a new proximity link with this transformation is added to the graph between the new node and the nearest one in the group. 
\end{enumerate}

\subsection{Graph Optimization Module}\label{sec:graphoptimization}
TORO (Tree-based netwORk Optimizer) \citep{Grisetti07a} is used for graph optimization. 
When loop closure and proximity links are added, the errors derived from odometry can be propagated to all links, thus correcting the local map. 
This also guarantees that nodes belonging to different maps are transformed into the same referential when loop closures are found. 

When only one map exists, it is relatively straightforward to use TORO to create a tree because it only has one root. 
However, for multi-session mapping, each map has it own root with its own reference frame. 
When loop closures occur between the maps, TORO cannot optimize the graph if there are multiple roots. 
It may also be difficult to find a unique root when some of the nodes have been transferred in LTM. 
As a solution, our approach takes the root of the tree to be the latest node added to the local map, which is always uniquely defined across intra-session and inter-session mapping. 
All other poses in the graph are then optimized using the last odometry pose as the referential.

\subsection{Path Planning Modules}
\label{planning}
Memory management has a significant effect on how to do path planning online using graph-based SLAM, for which the map changes almost at each iteration and with only the local map accessible while executing the plan. 
This differs from approaches that assume that the map is static and/or that all the previously visited locations always remain in the map. 
In this paper, SPLAM-MM uses two path planners: a Metrical Path Planner (MPP) and a Topological Path Planner (TPP).

\subsubsection{Metrical Path Planning Module}
\label{sec:mpp}

\begin{figure*}[!t]
\centering
\begin{tabular}{ccc}
\subfloat[]{\includegraphics[width=1.9in]{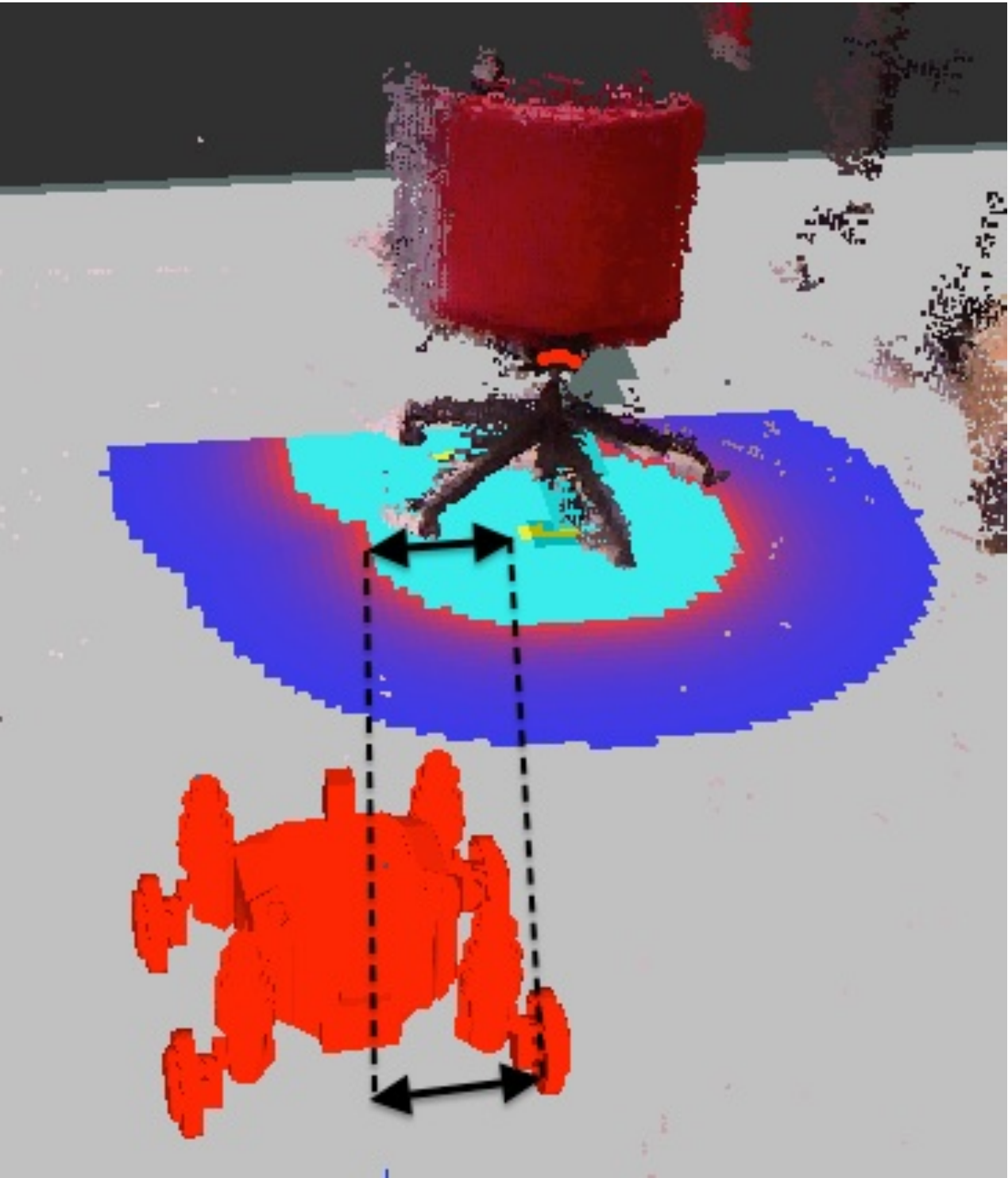}\label{fig:chairA}} &
\subfloat[]{\includegraphics[width=1.9in]{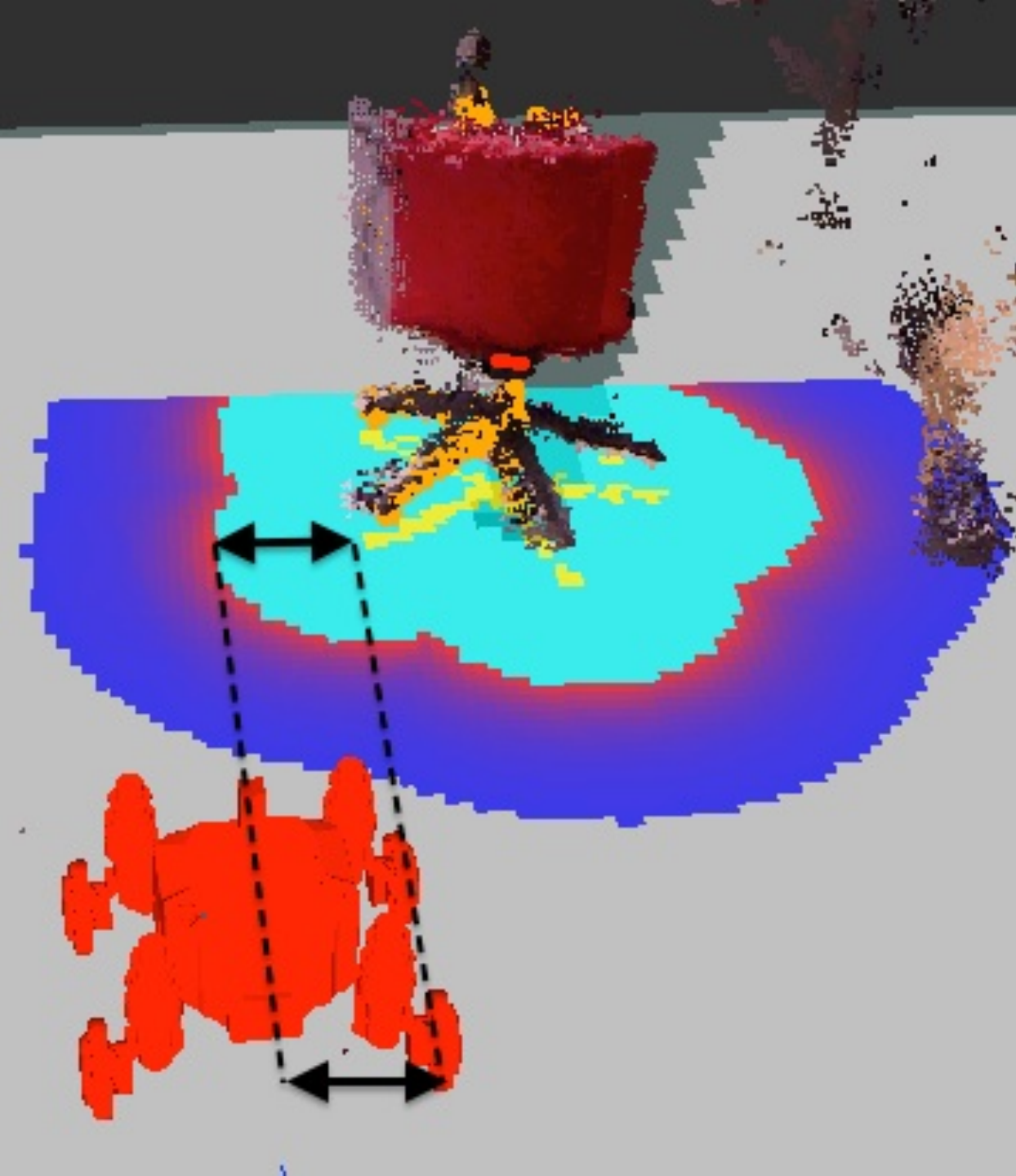}\label{fig:chairB}} &
\subfloat[]{\includegraphics[width=2.2in]{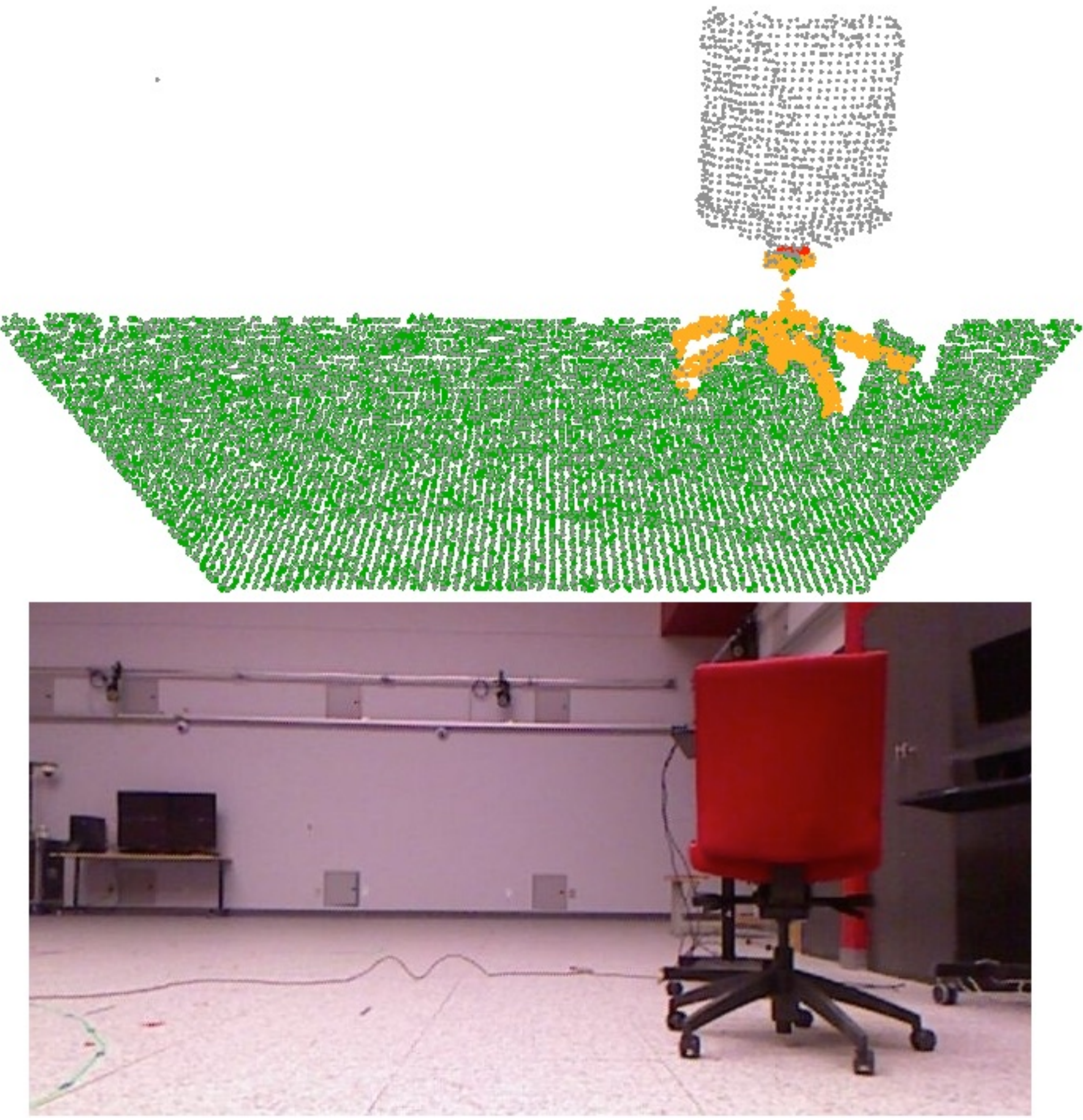}\label{fig:chairC}}
\end{tabular}
\caption{Example of obstacle detection using the laser rangefinder and the RGB-D camera. 
The red dots on the chair show what is detected using the laser rangefinder data.
The cyan area is derived from the obstacle projection on the ground plane up to robot's footprint radius, delimiting where the center of the robot should not enter to avoid collisions.  
In a), only the laser rangefinder data are used and the chair's wheels are not detected, making unsafe for the robot to plan a path around the chair. In b), the point cloud generated from the camera's depth image is used and the chair's wheels are detected (shown by the orange dots), increasing the cyan area (and consequently the area to avoid colliding with the chair). 
Illustration c) presents a view from the RGB-D camera where the segmented ground is shown in green and the obstacles in orange.}
\label{fig:example_chair}
\end{figure*}

MPP receives a pose expressed in ($x, y, \theta$) coordinates, and uses the local map to plan a trajectory and to make the robot move toward the targeted pose while avoiding obstacles.
Our MPP implementation exploits the ROS navigation stack \citep{officemarathon}  to compute trajectories expressed as a sequence of velocity commands (expressed as twists) sent to the robot's Motion Controller module. 
A global Costmap is used to plan a trajectory to a targeted pose. 
MPP creates the global Costmap from an occupancy grid created using the assembled laser scans from the latest local map. 
Each time the local map is updated, the occupancy grid is re-assembled and the trajectory is re-planned. 
MPP also uses a local Costmap for its Dynamic Window Approach (DWA) \citep{fox1997dynamic} to handle dynamic obstacles for collision avoidance. 
The local Costmap is created directly from sensor readings.  
To create the local Costmap, only using the laser rangefinder for obstacle detection revealed to be insufficient:
while the laser range finder can detect most of the obstacles (e.g., walls, people, table legs), it is located 40 cm above the floor and all obstacles under this height cannot be detected. 
Therefore, the depth image from the RGB-D camera is also used to detect these small obstacles and to add them to the local Costmap. 
Figure \ref{fig:example_chair} shows an example where combining laser scans and RGB-D data creates a more robust and a safer local Costmap for navigation.
Note that segmentation of the point cloud generated from the depth image is required to be able to add or clear small dynamic obstacles below the RGB-D camera. 
To segment the ground, all points with normal parallel to $z$-axis (up to an angle $Z$) are labeled as ground. 
Then, all other points under a maximum height $U$ are labeled as obstacles. 
This method would also make the robot capable of operating on uneven terrain. 

\subsubsection{Topological Path Planning Module}
\label{sec:tpp}

When TPP receives a goal identified by a node ID from a user (or a high-level module like a task planner, or in this paper the Patrol module), the global map is provided by the graph-based SLAM-MM module, and a topological path is computed to reach this goal. 
The topological path is a sequence of poses, expressed by their respective node IDs, to reach the goal. 
This step must be done offline or when the robot is not moving because all nodes linked to the current local map should be retrieved from LTM to build the global map. 

To choose which nodes to use for navigation, TPP computes a path from the current node to the goal node using Djikstra algorithm \citep{dijkstra1959note}. 
The choice of using Dijkstra over A* is to avoid global graph optimization, which is time consuming, to know the distance to goal required by A*. 
Dijkstra can also be computed directly when fetching the global map from LTM.  
Similar to \citep{valencia2013planning}, to avoid losing track of the planned path, TPP prefers paths traversed in the same direction (e.g., where the camera is facing the same direction than on the nodes on the path) over shortest paths. 
This increases localization confidence: loop closure detection and visual proximity detection are more reliable than proximity detection using only laser scans because of their double verification (3D visual word correspondences and 2D ICP transformation refinement). 
To embed this preference in Djikstra, the search cost is angular-based instead of distance-based, i.e., it finds the path with less orientation changes when traversing it in the forward direction. 

Then, TPP selects the farthest node on the path in the local map and sends its pose to MPP.
While MPP makes the robot navigate to its targeted pose, TPP indicates to the graph-based SLAM-MM module which upcoming nodes on the topological path is needed, expressed as a list of node IDs from the latest node reached on the path to the farthest node inside the radius $R$ (to limit the size of the list). 
The required nodes are identified by the graph-based SLAM-MM module with Heuristic 2 either to remain in WM or to be retrieved from LTM to extend the local map. 
The maximum number of retrieved nodes per map update is limited to $M$ because this operation is time consuming as it needs to load nodes from LTM. 
$M$ is set based on the hardware on which LTM is saved and according to the maximum velocity of the robot: for instance, if the robot is moving at the same speed or less as when it traversed the same area the first time, $M=1$ would suffice to retrieve nodes on the path without having to slow down to wait for nodes not yet retrieved. 

Extending the local map with nodes of the topological path is important for the robot to localize itself using the Appearance-based Loop Closure Detection module or using the Proximity Detection module, making it able to follow the topological path appropriately. 
As the robot moves and new local maps are created, TPP always looks for the farthest node of the topological path that can be reached in the local map to update the current pose sent to MPP module. 
If new nodes are retrieved from LTM on the topological path, then the farthest pose is sent to MPP. 
TPP also detects changes in the local map after graph optimization (e.g., when new loop closures are detected): if so, the updated position of the current pose is sent to MPP.

Up to a ratio $O$ of the WM size, nodes identified by the planner and located in the radius $R$ from the robot's current position are immunized to be transferred, with $R$ being the sensor range.

\begin{figure*}[!t]
\centering
\begin{tabular}{ccc}
\subfloat[]{\includegraphics[width=2in]{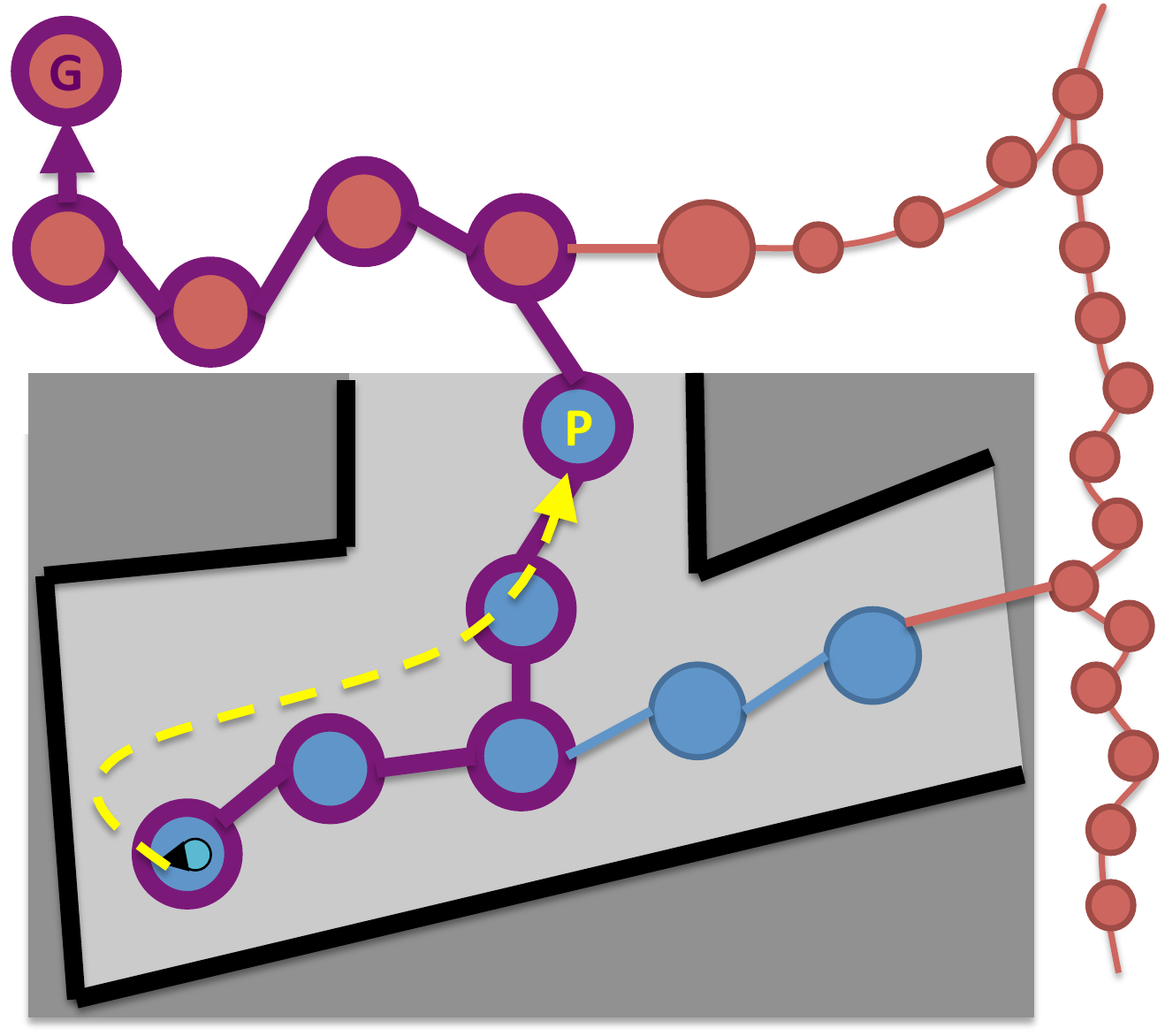}\label{fig:example_planner_a}} &
\subfloat[]{\includegraphics[width=2in]{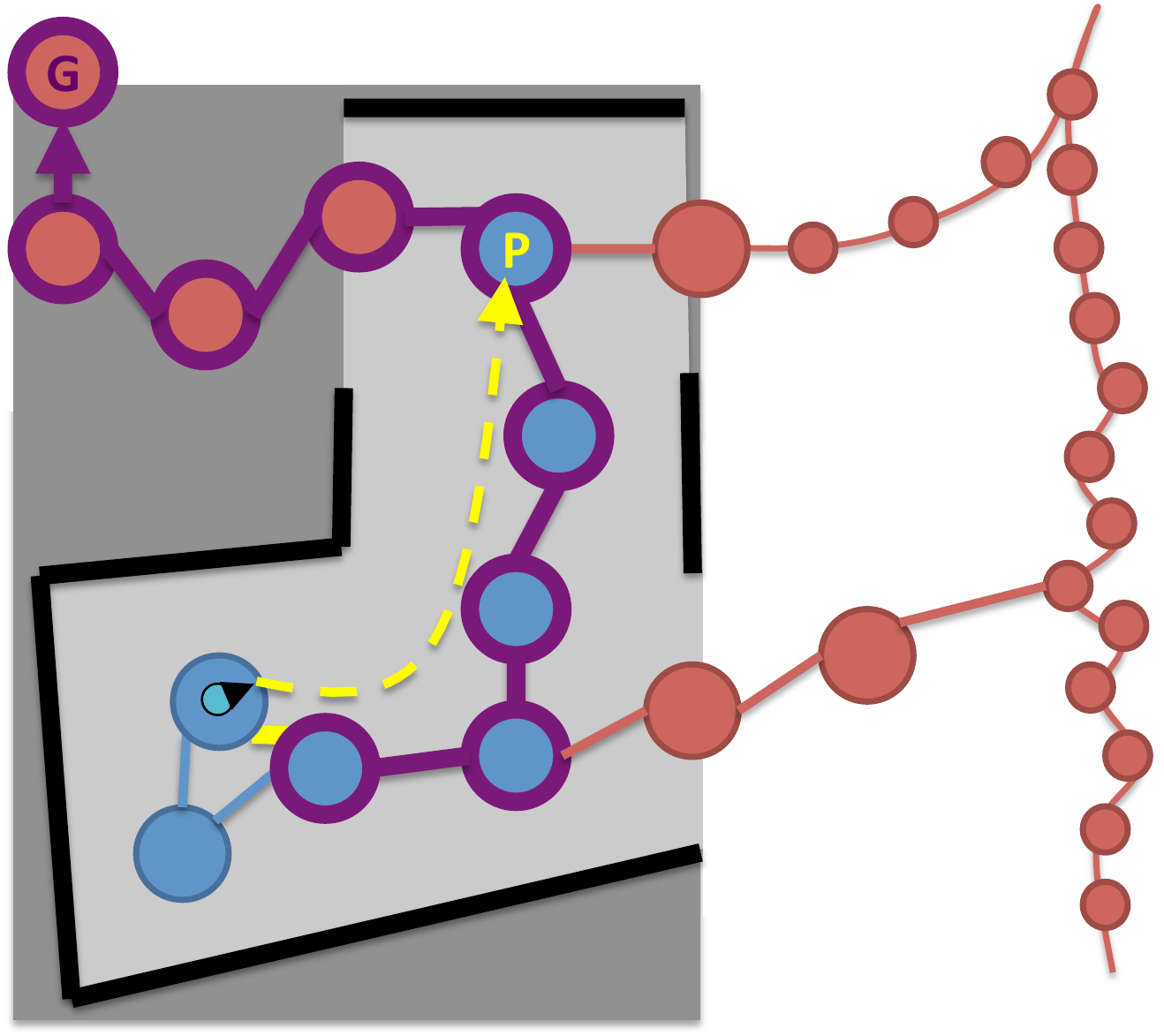}\label{fig:example_planner_b}} &
\subfloat[]{\includegraphics[width=2in]{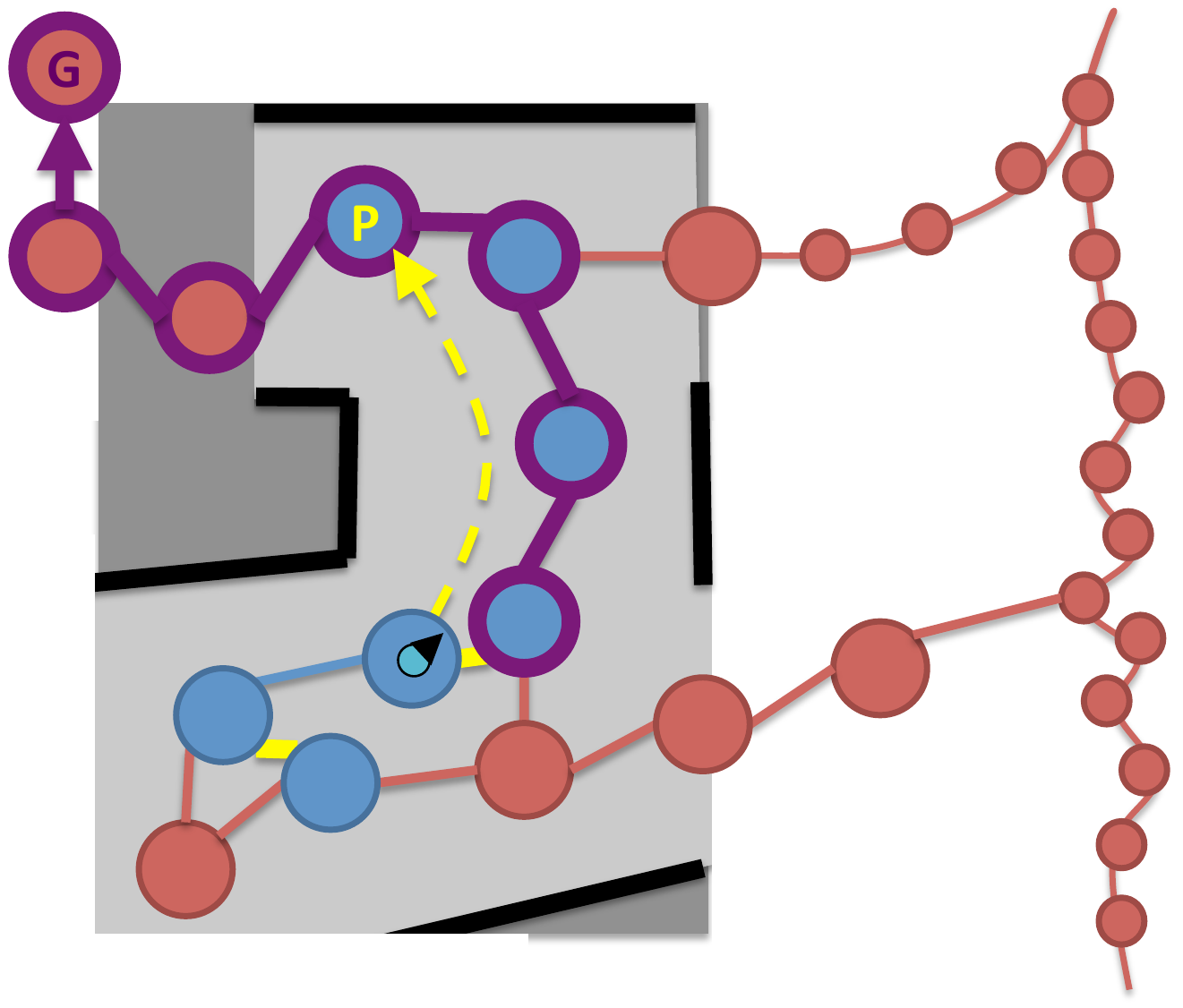}\label{fig:example_planner_c}} \\
\subfloat[]{\includegraphics[width=2in]{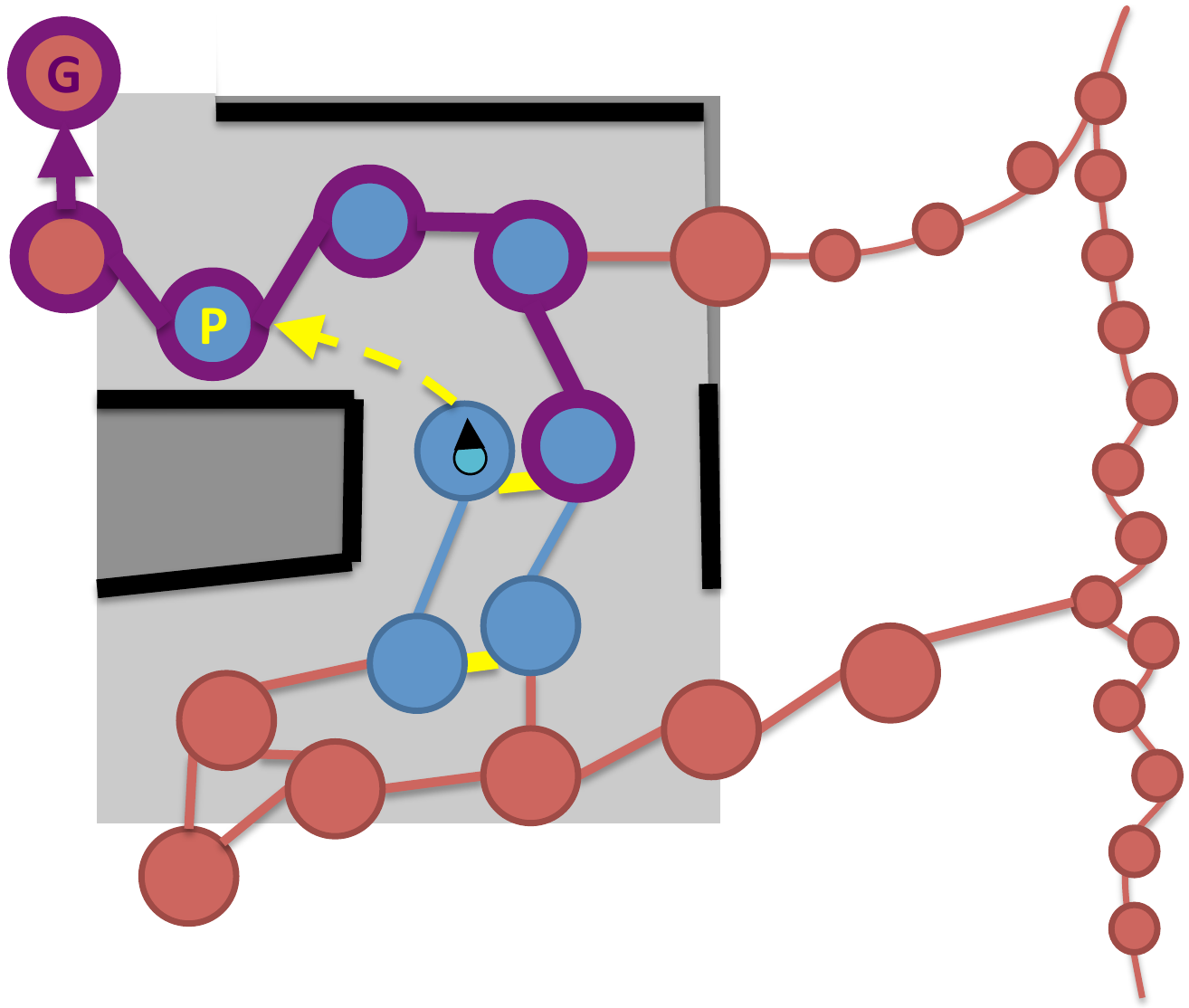}\label{fig:example_planner_d}} &
\subfloat[]{\includegraphics[width=2in]{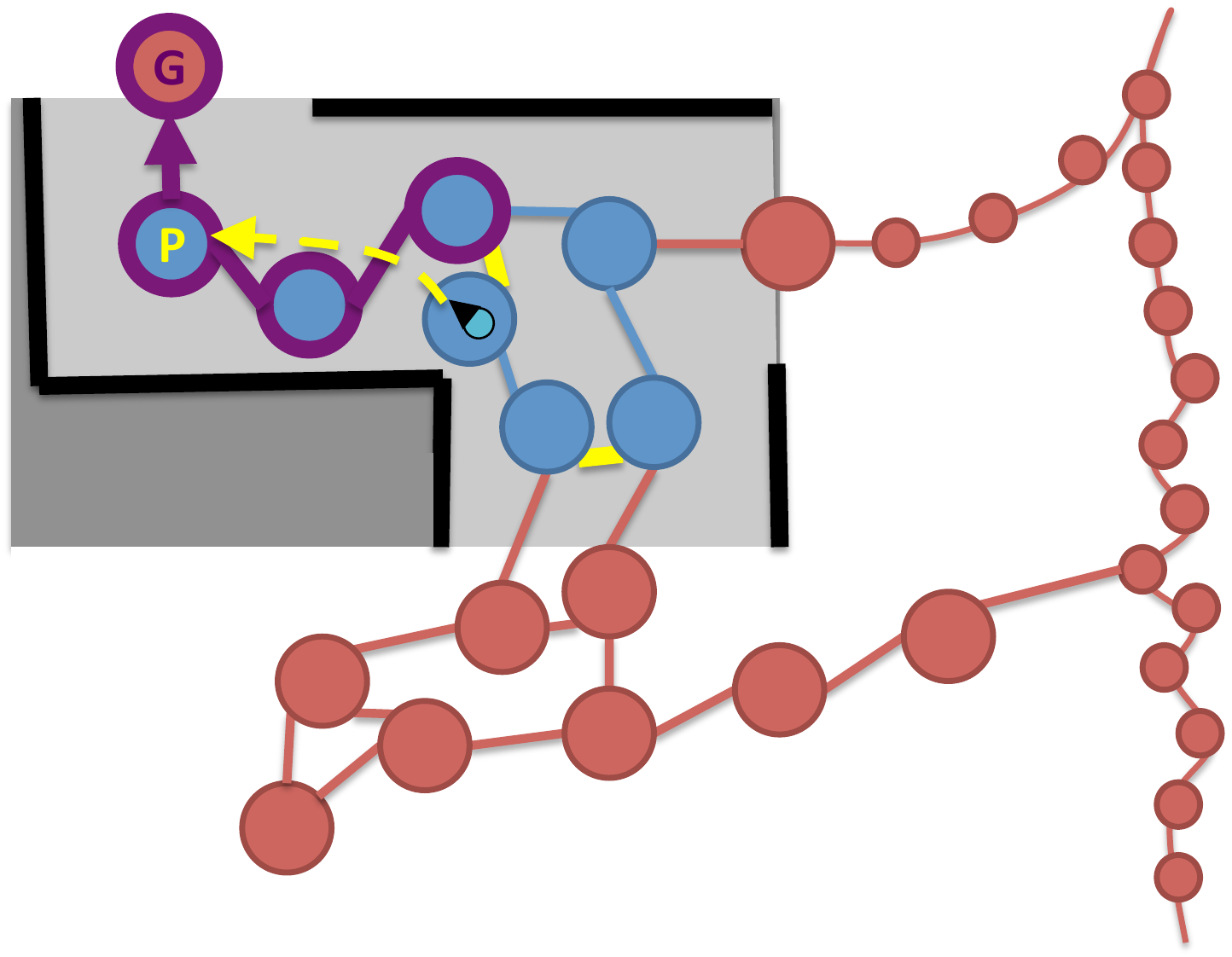}\label{fig:example_planner_e}} &
\subfloat[]{\includegraphics[width=2in]{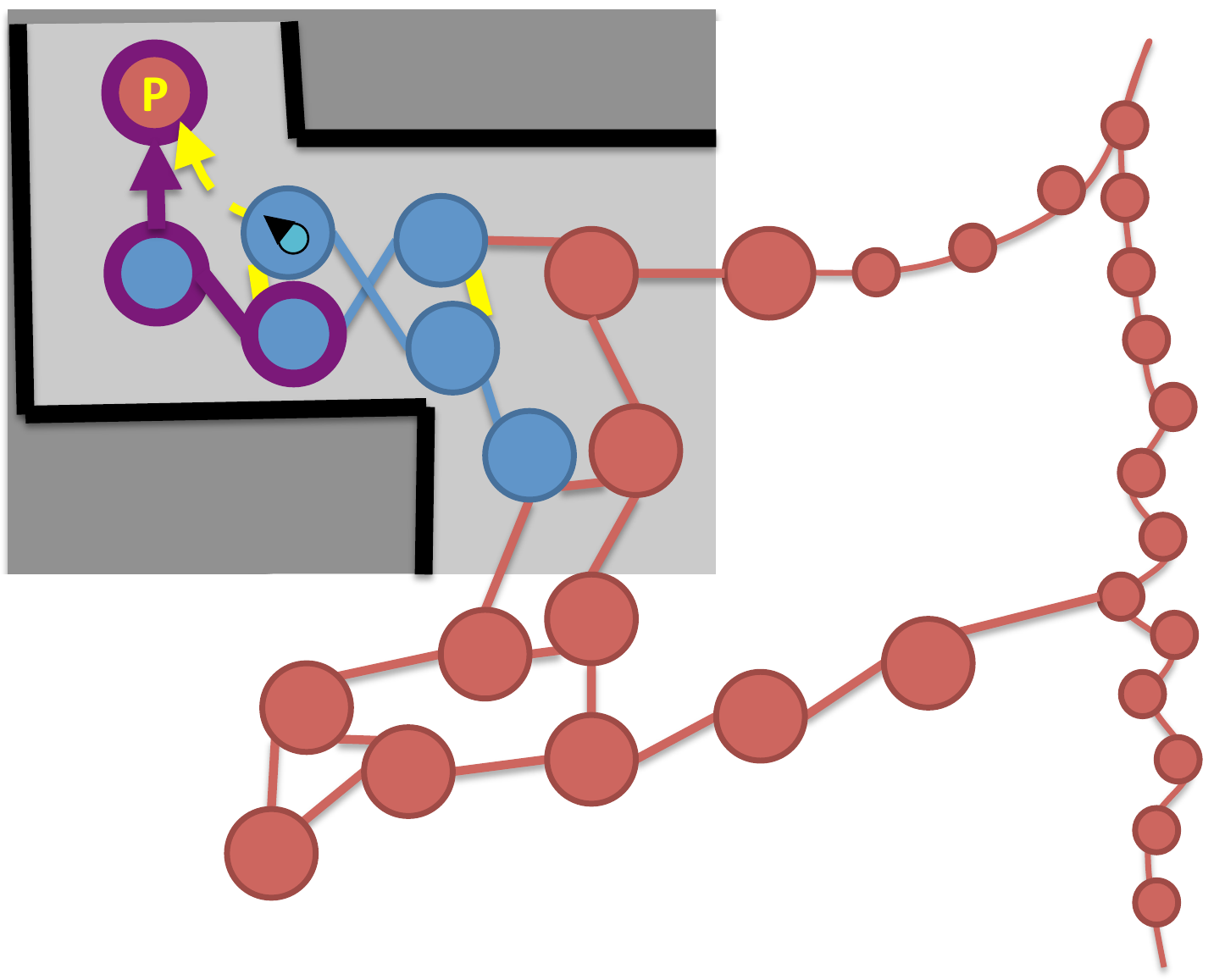}\label{fig:example_planner_f}} \\
\end{tabular}
\caption{Interaction between TPP and MPP for path planning. The goal is identified by the purple G. The topological path is shown with purple links. The dashed yellow arrow is the trajectory computed by MPP to the targeted poses designated by the yellow P. Light gray, dark gray and black areas of the occupancy grid represent free, unknown and occupied cells, respectively. Blue nodes are in WM, and red nodes are in LTM. Yellow links are proximity links. 
}
\label{fig:example_planner}
\end{figure*}

Figure \ref{fig:example_planner} presents an example of the interaction between MPP and TPP to reach a goal G.
While the robot is moving, TPP always sends the farthest pose P of the node on the topological path (purple links) in the local map. 
An occupancy grid is assembled with the laser scans contained in the nodes of the local map. MPP uses this occupancy grid to plan a trajectory (yellow arrow) to P. 
To keep the WM size constant, as nodes are retrieved from LTM on the path, older nodes are transferred to LTM. 
To follow the path appropriately, proximity links are detected to correct the map as the robot moves, otherwise the situation explained by Fig. \ref{fig:hallproblem}a would happen.

TPP iterates by sending poses until the node of the goal (under a goal radius $D$ expressed in m) is reached.
Finally, handling situations where the environment has changed too much for proper localization must be taken into consideration. If no loop closures and proximity detections occur when following a path, a temporary link is added between the current node and the closest one in the path so that the topological path is always linked to the current node in the local map. 
Without this link, if previous nodes between the current node and those of the topological path are transferred to LTM, the local map would be divided and the nodes of the path would not be in the local map anymore. 
This temporary link is removed when a new link is added between the current node and the closest one in the path or when the goal is reached. 
If the robot has not reached the current pose set to MPP after $F$ iterations of SPLAM-MM (e.g., MPP cannot plan to the 
requested pose because of the presence of a new obstacle or because the robot cannot localize itself on the path), TPP chooses another pose on the upcoming nodes and sends it to MPP. 
If all the upcoming nodes cannot be reached, TPP fails and sends a status message to its connected modules so that they can be notified that the goal cannot be reached. 

\subsection{Patrol Module}
We implemented the Patrol module to generate navigation goals, referred to as waypoints so that the robot is programmed to continuously patrol an area. 
The Patrol module receives waypoints as inputs and sends them successively to TPP. 
By examining TPP's status messages, Patrol can know when a goal is reached or if TPP has failed. 
Whenever the status indicates that the goal is reached or not, the Patrol module sends the next waypoint, and restart to the first one once the whole list has been processed.

\section{Results}
\label{sec:results}
\label{results}

Table \ref{table_parameters} shows the parameters used for the trials\footnote{In comparison with \citep{labbe13appearance}, $T=T_{time}$, $S=T_{STM}$ and $Y=T_{similarity}$.}. 
The acquisition time $A$ used is 1 sec (i.e., the map update rate is 1 Hz), which set the maximum online time allowed to process each node added to the map. 
For the trials, $T$ is set to 200 ms to limit CPU usage for SPLAM-MM to around 20\%, to make sure that higher frequency modules (acquisition of Sensor Data acquisition and MPP) 
can run at their fixed frequency of 10 Hz. 
The robot is relatively moving at the same velocity during the trials, and therefore $M$ is fixed to 2 to make sure that nodes on a planned path are retrieved fast enough to avoid having the robot wait for nodes still in LTM. All computations are done onboard on the robot, which is equipped with a 2.66 GHz Intel Core i7-620M and a 128 GB SSD hard drive (on which the LTM is saved). 

\begin{table}[!t]
\renewcommand{\arraystretch}{1.3}
\caption{Parameters used for the trials}
\label{table_parameters}
\centering
\begin{tabular}{lcc}
\hline
Acquisition time                        			& $A$     	& $1$ sec \\ 
ICP correspondence ratio                		& $C$     	& 0.3 \\
Radius of the goal area                     		& $D$     	& 0.5 m \\
TPP iterations before failure                	& $F$     	& 10\\
Loop closure hypothesis threshold       	& $H$     	& 0.11 \\ 
Minimum RANSAC visual word inliers      	& $I$     	& 5 \\ 
Close nodes radius                      		& $L$     	& 0.5 m \\
Maximum retrieved close nodes           	& $M$     	& 2 \\  
Heuristics 2 close-by nodes ratio          	& $O$     	& 0.25 \\
Laser scan range                        		& $R$     	& 4 m\\
STM size                                			& $S$     	& 20 \\ 
Time limit                              			& $T$     	& 200 ms \\
Maximum obstacle height           		&  $U$	& 0.4 m \\
Similarity threshold                    			& $Y$     	& 0.3 \\
Ground segmentation maximum angle   	& $Z$     	& $0.1$ rad \\ 
\hline
\end{tabular}
\end{table}

\begin{figure}[!t] 
\centering 
\includegraphics[width= \columnwidth]{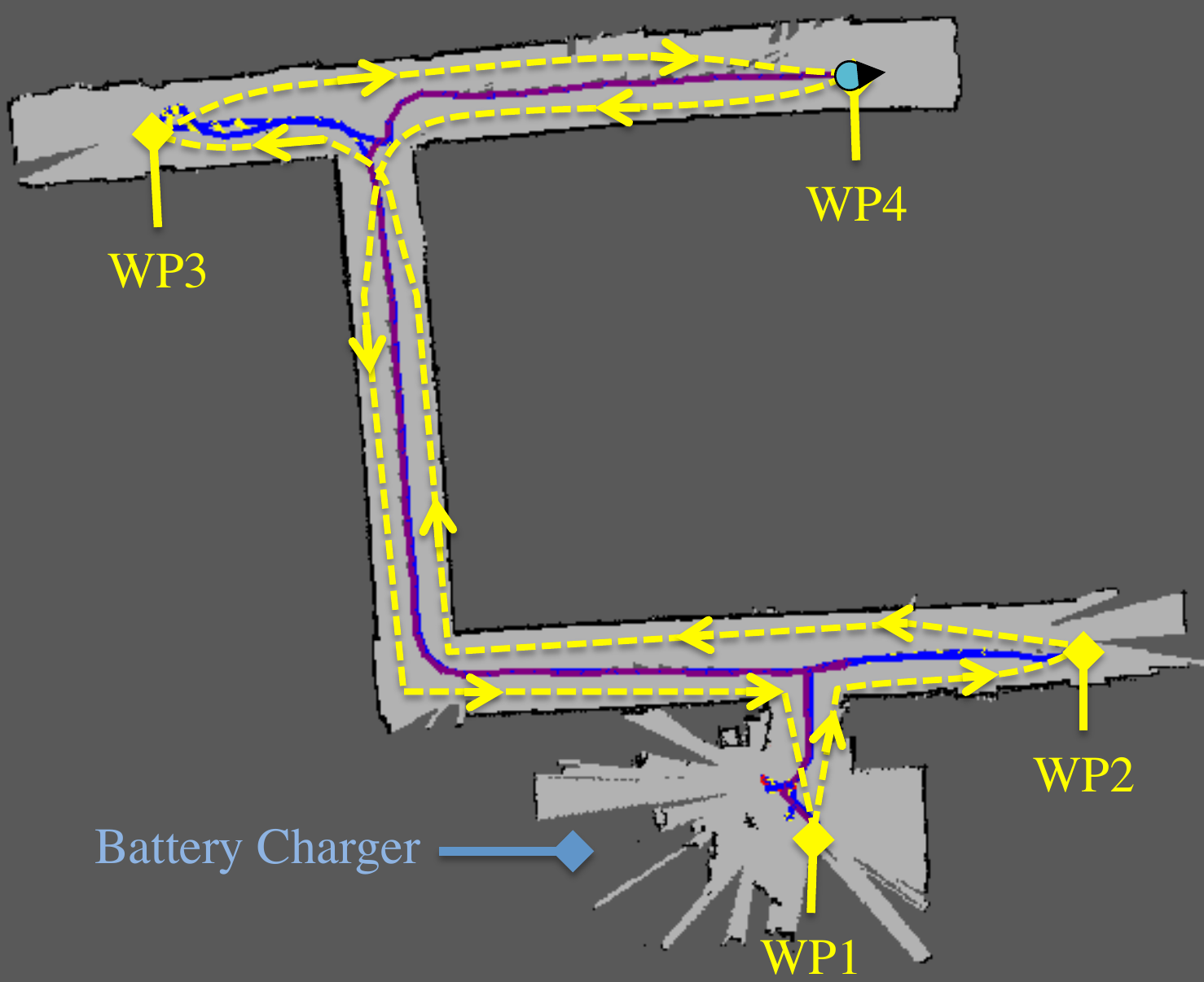} 
\caption{Waypoints WP1 to WP4 identified on the global map. The purple path is the first path planned by TPP from the WP4 to WP1.} 
\label{fig:waypoints} 
\end{figure}

\begin{figure}[!ht]
\centering
\begin{tabular}{cc}
\subfloat[]{\includegraphics[width=1.5in]{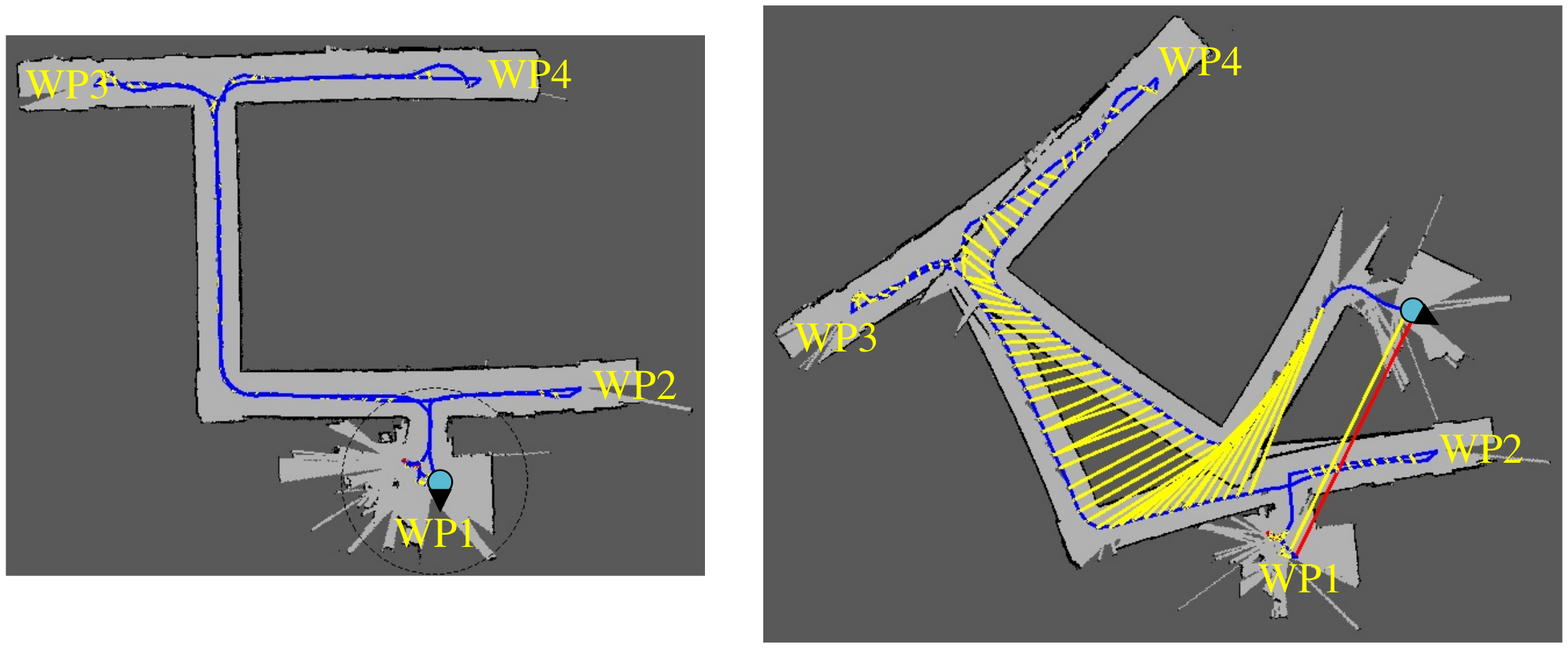}\label{fig:first_map_optimized}} &
\subfloat[]{\includegraphics[width=1.5in]{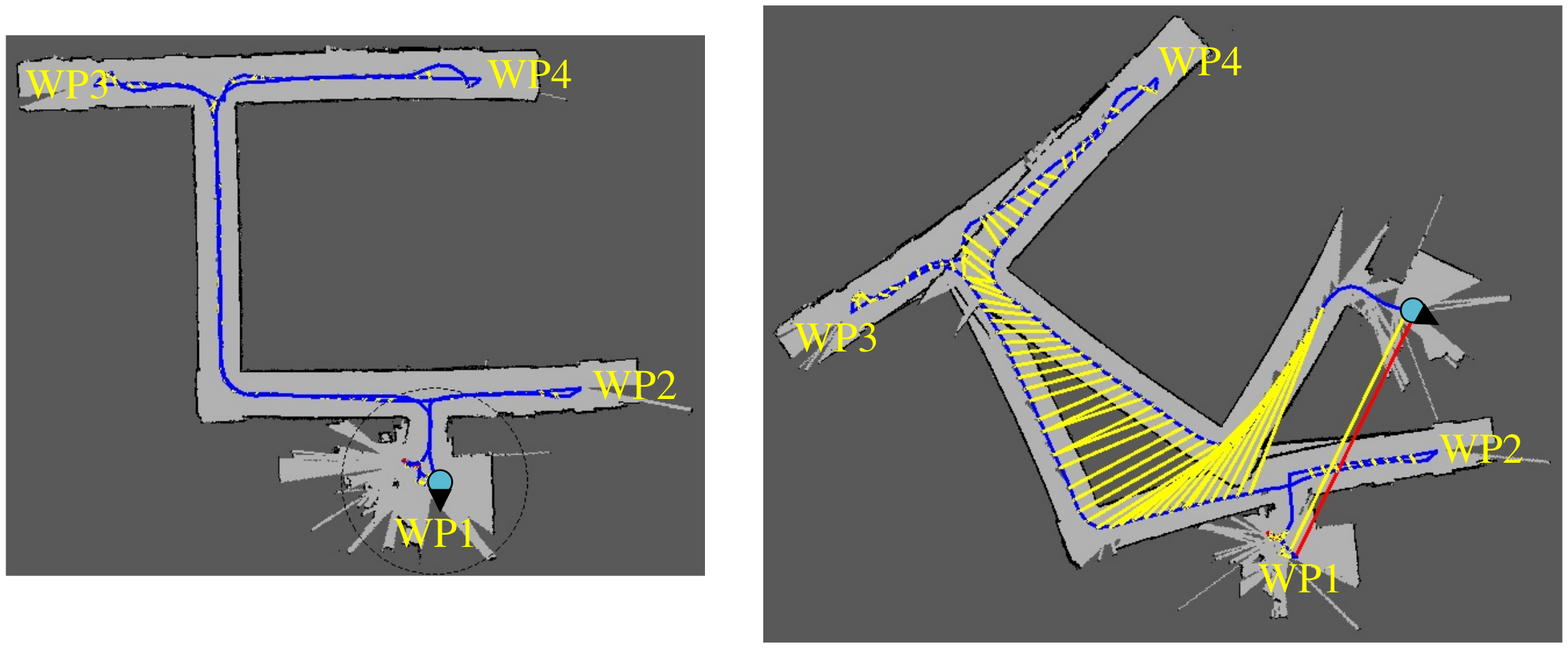}\label{fig:first_map_raw}} 
\end{tabular}
\caption{Global maps, optimized and not optimized, after reaching WP1. Yellow and red links are proximity and loop closure links, respectively.}
\label{fig:first_maps}
\end{figure}

To define the area over which the robot had to patrol, during session 1 we first teleoperated the robot and defined four waypoints (WP1 to WP4). 
There were no people in the environment during the teleoperation phase. 
After reaching WP4, the autonomous navigation phase is initiated by sending the waypoints to the Patrol module. 
Figure \ref{fig:waypoints} illustrates the four waypoints on the global map and the first planned trajectory by TPP (purple path) from the current position of the robot (WP4) to WP1. 
To come back to WP1, the robot had to follow the path in the opposite direction from when these nodes were created. 
Proximity detection made it able to follow the path appropriately. 
To see more clearly the effect of proximity links, Fig. \ref{fig:first_maps} shows the maps after reaching WP1 with and without graph optimization. 
Navigation would not have been possible without proximity links: the local map would have look like the map in (b) without the yellow links because no appearance-based similarities would have been found with nodes from the map on the planned path. 
When reaching WP1, the Patrol module sends the next waypoint (WP2), making the robot continue patrolling. 

\begin{figure*}[!t] 
\centering 
\includegraphics[width= \textwidth]{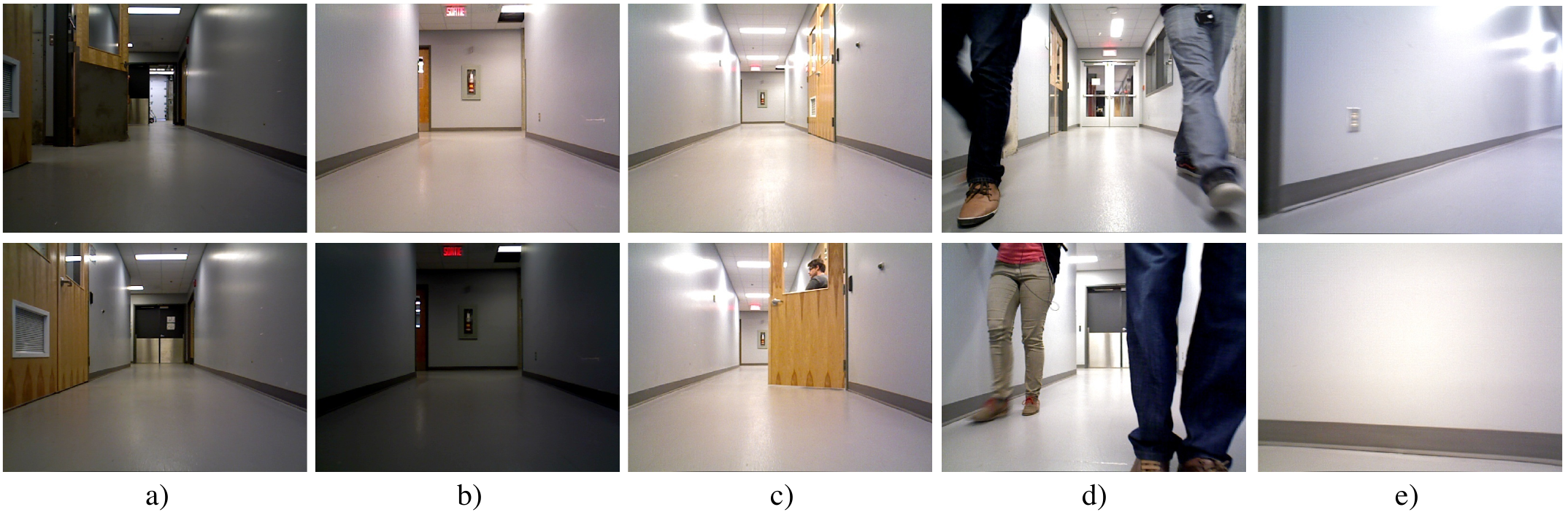} 
\caption{Events that occurred during the trials: a) open and closed doors between traversals; b) camera exposure that led to the extraction of different visual features, making it difficult to find loop closures; c) someone opening a door while the robot is navigating; d) people walking around or blocking the robot; e) featureless images on which loop closure detection cannot work.} 
\label{fig:obstacles} 
\end{figure*}

Every 45 minutes or so of operation, the robot was manually shutdown and moved to the battery charger near WP1. 
Once recharged, a new session of SPLAM-MM was initiated, creating a new node in STM with odometry reset, while preserving the nodes in WM and LTM. 
As the robot was initialized in the area of WP1 for each session, loop closures were found, connecting and optimizing the new map with nodes created from previous sessions, and allowing the Patrol module to provide waypoints as navigation goals to patrol the area.  
Overall, 11 indoor mapping sessions were conducted, for a total distance of $10.5$ km lasting 7.5 hours of operation spent over two weeks. 
The robot did 111 patrolling cycles (i.e., traversing from WP1 through WP2, WP3, WP4 and coming back to WP1).  
The sessions were conducted during office hours, with people walking by. 
A total of 139 people were encountered by the robot while patrolling. 
Figure \ref{fig:obstacles} illustrates the dynamic conditions and some of the obstacles that the robot had to deal with during the trials. 

The main goal of the trials is to see how SPLAM is influenced by memory management over long-term operation, only having the local map for online processing. 
This can be illustrated by looking at the influences of memory management on SPLAM, interactions between TPP and MPP, and the influences of LTM on TPP.
As the robot is continuously adding new nodes, the trials also demonstrate how SPLAM-MM works in an unbounded environment. 

\subsection{Influences of MM on SPLAM}

\begin{figure*}[!t] 
\centering 
\includegraphics[width= \textwidth]{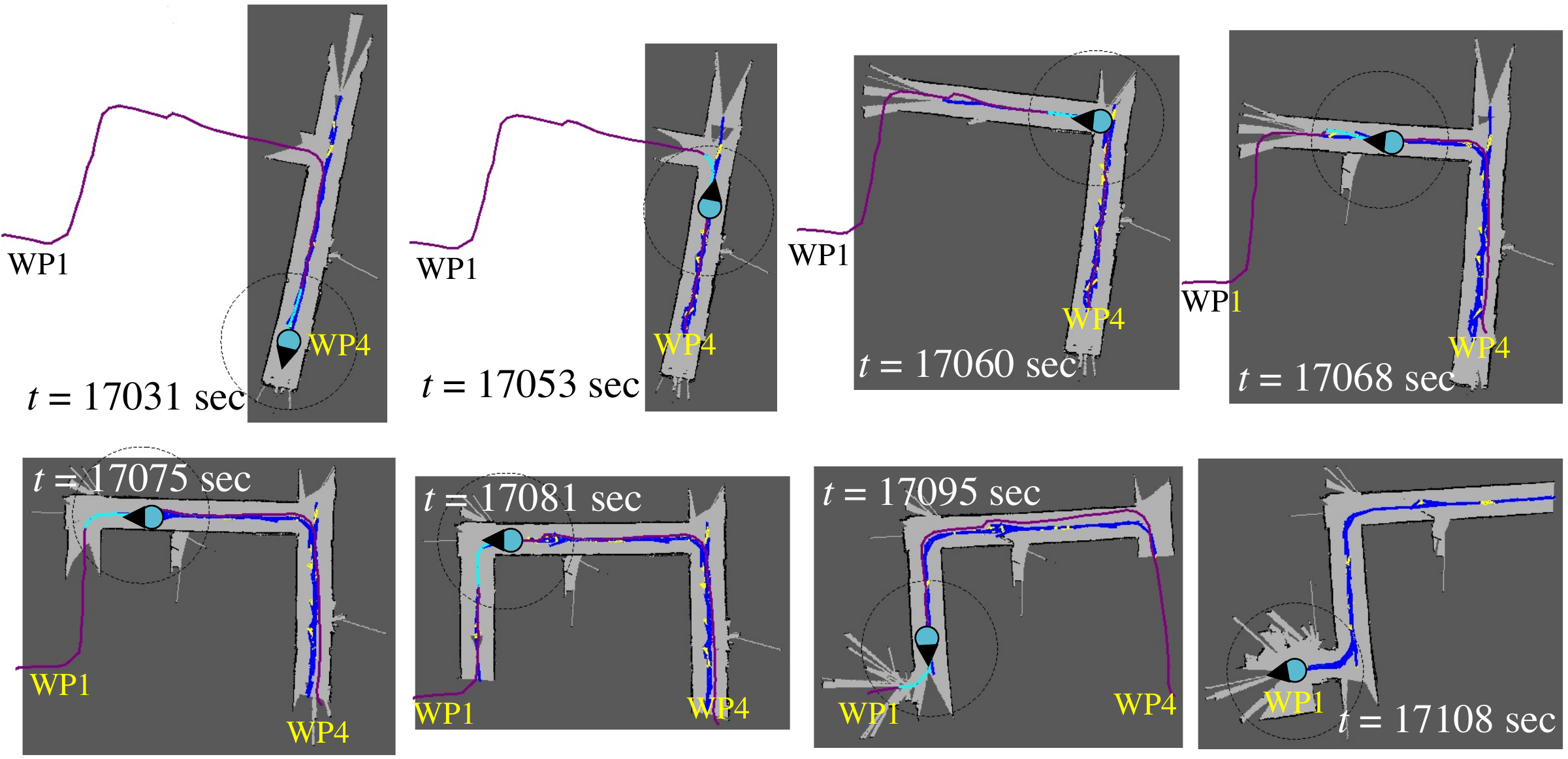} 
\caption{Example of the effect of memory management when travelling from WP4 to WP1 after 4.7 hours of operation. The path planned is shown in purple. The small colored icon represents the robot position at each time step. The dotted circle around the robot position illustrates the laser scan range $R$. The cyan lines represent the upcoming nodes on the planned path.} 
\label{fig:planningResults} 
\end{figure*}

Figure \ref{fig:planningResults} shows a typical navigation result when reaching the time limit $T$, thus limiting the size of the local map used for online navigation.  
This example shows the path planned between WP4 and WP1 after 4.7 hours of operation. 
The local maps used for online planning, localization and mapping are shown for different time steps along the trajectory.
At $t=17031$ sec, the planned path had 67 nodes and was 33 m long.
It took $1.3$ sec to be generated by TPP and to have the first pose on the path sent to MPP. 
The laser scan range $R$ is delimiting the upcoming nodes on the path provided by TPP.
As the robot navigates in the environment, the farthest available pose in the local map on the path (end of the cyan line) is sent from TPP to MPP. 
Upcoming nodes, if they are not in WM, are retrieved to make the robot able to localize itself (though loop closures and proximity detections) on the path. 
Looking at how the local map changes in these snapshots, notice how starting from $t=17075$ sec, the initial portion of the path is transferred in LTM to keep the size of the WM relatively constant. 
At $t=17108$ sec, the robot reached WP1. 

\begin{figure*}[!t] 
\centering 
\includegraphics[width= 5.5in]{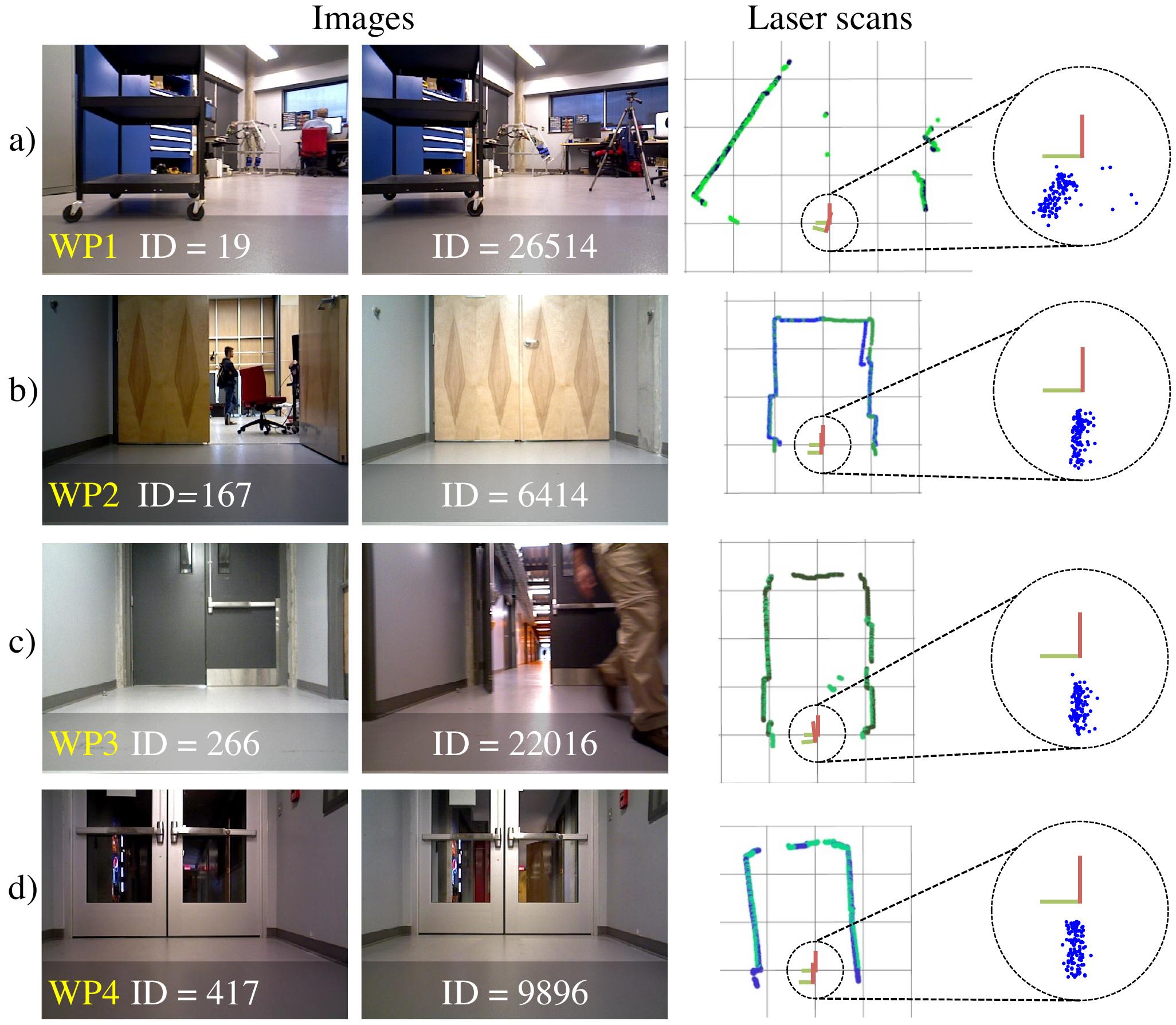} 
\caption{Comparison of the corresponding images between the waypoint (left image) and at the last pose reached on one of the planned path (right image) for the waypoints. The top view grid shows the laser scan readings and referentials of the waypoint's nodes (at the origin of the grid) and the final node.  
The zoomed portions represent the final poses of the robot (represented by blue dots), for all paths planned for each waypoint. The circle represents the goal radius $D$, and the grid's cells used for visualization have a width of 1 m.} 
\label{fig:goals} 
\end{figure*}

Figure \ref{fig:goals} compares the images between each waypoint and the final position of the robot at the waypoints. 
The robot successfully reached the waypoints (within $D$ as the goal radius) 445 out of 446 times. 
For WP2, WP3 and WP4, the robot always came from behind the waypoint, and as soon the robot reached the waypoint within a $D$ radius, TPP detected that the goal was reached. 
This explains why all the poses are behind the waypoints but inside the goal radius $D$. 
Similarly, for WP1, the robot came from behind from a slightly different direction. 
Spurious poses on the right part of the circle are those where there was an obstacle that caused the robot to avoid it, making it reach the waypoint from a different direction. 
The one time the robot failed to reach a waypoint is because someone blocked the robot for a long time, making TPP failed after $F$ attempts of reaching the upcoming nodes: a failure status message was then sent to the Patrol module to provide the next waypoint. 
The person left soon after the next waypoint was sent, and the robot reached the new waypoint provided. 

\begin{figure*}[!ht]
\centering
\begin{tabular}{cc}
\subfloat[Number of nodes in WM and in the local map.]{\includegraphics[width=\columnwidth]{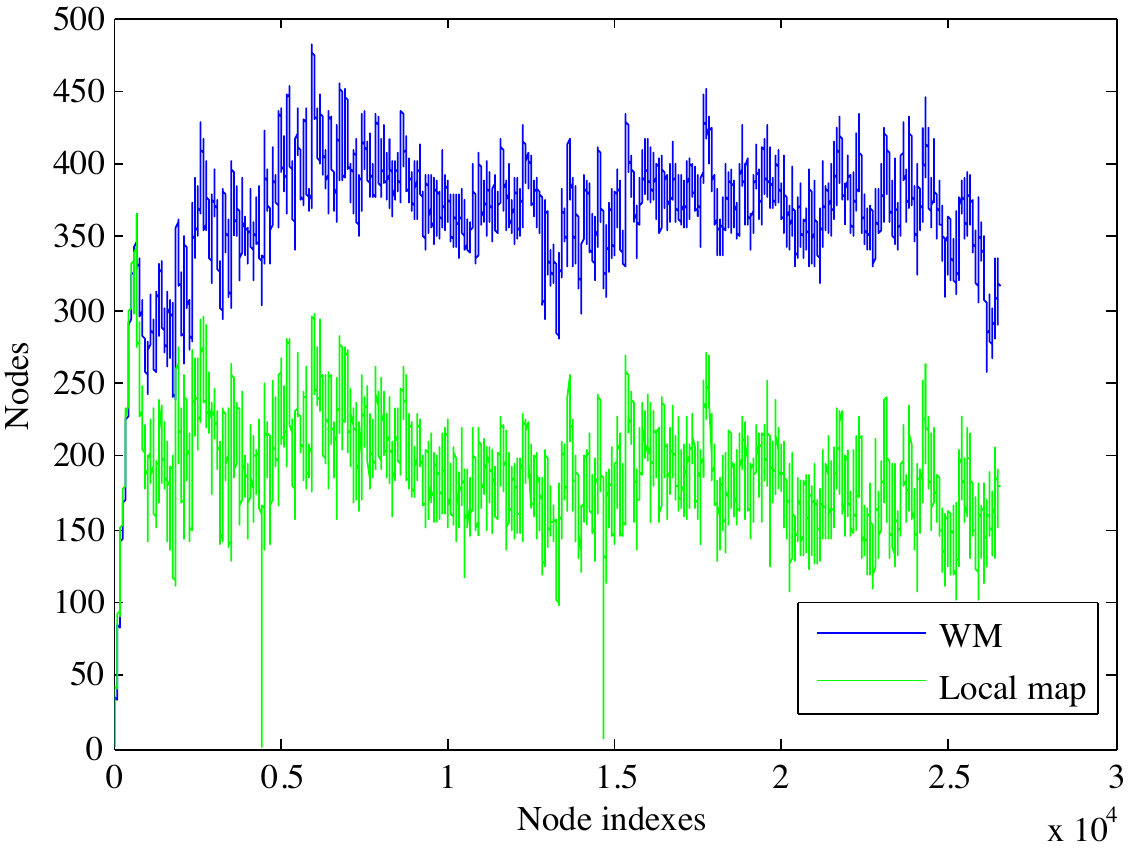}\label{fig:wm_local_graph}} &
\subfloat[Processing time (the horizontal line represents $T=0.2$ sec).]{\includegraphics[width=\columnwidth]{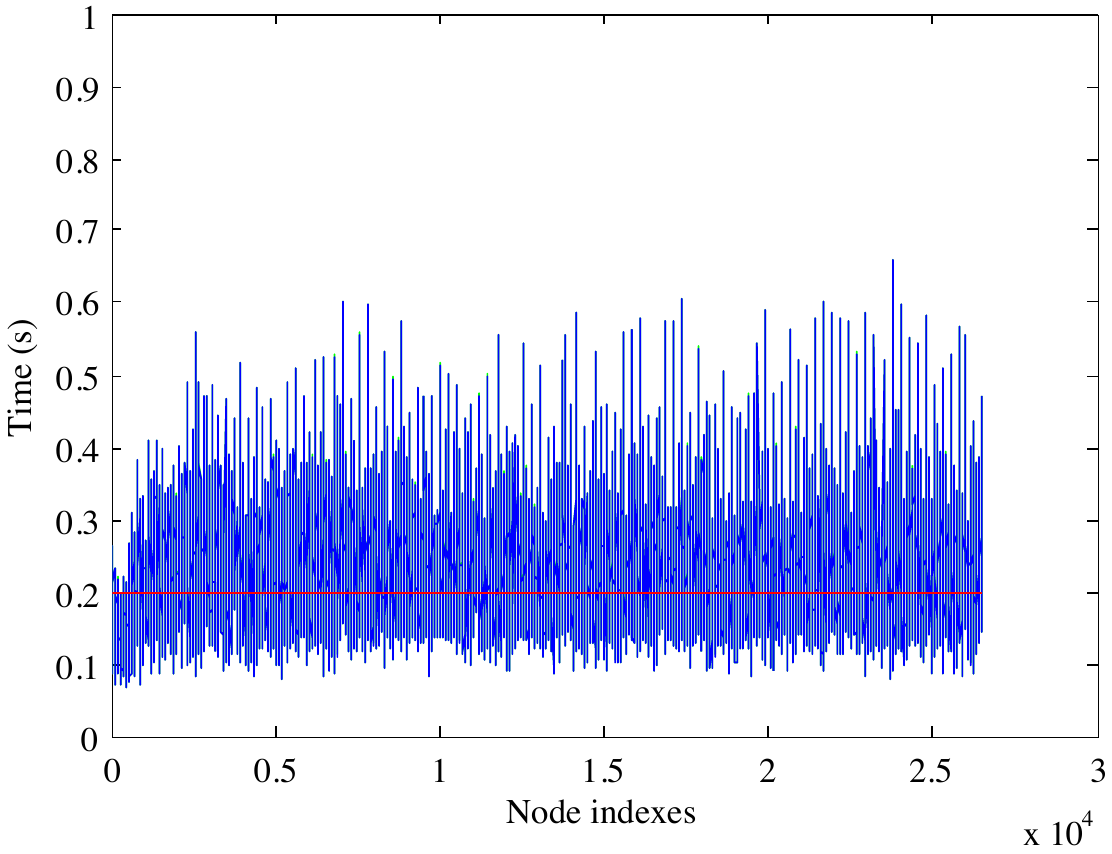}\label{fig:timeAll}} \\
\end{tabular}
\caption{Memory size and total processing time over the 11 mapping sessions.}
\label{fig:graphAndTime}
\end{figure*}

Figure \ref{fig:graphAndTime} illustrates the evolution of the number of nodes in WM and online processing time over the 11 mapping sessions. 
Processing time includes all SPLAM-MM modules except MPP which was running concurrently on a separate process (its processing time is only dependent of the local map size). 
As explained in Section \ref{sec:tpp}, TPP occurs offline and only when a new goal is received from the Patrol module, and is examined in Section \ref{sec:gr}.
Fig. \ref{fig:wm_local_graph} illustrates that the number of nodes in WM and the local map was identical until $T$ sec was reached. 
After that, nodes were transferred to LTM to limit the WM size for online processing, which is satisfied as shown by Fig. \ref{fig:timeAll}. 
Processing time also remained well under the acquisition time $A$.

\subsection{TPP-MPP Interactions}
\begin{figure}[!t]
\centering
\subfloat[]{\includegraphics[width=\columnwidth]{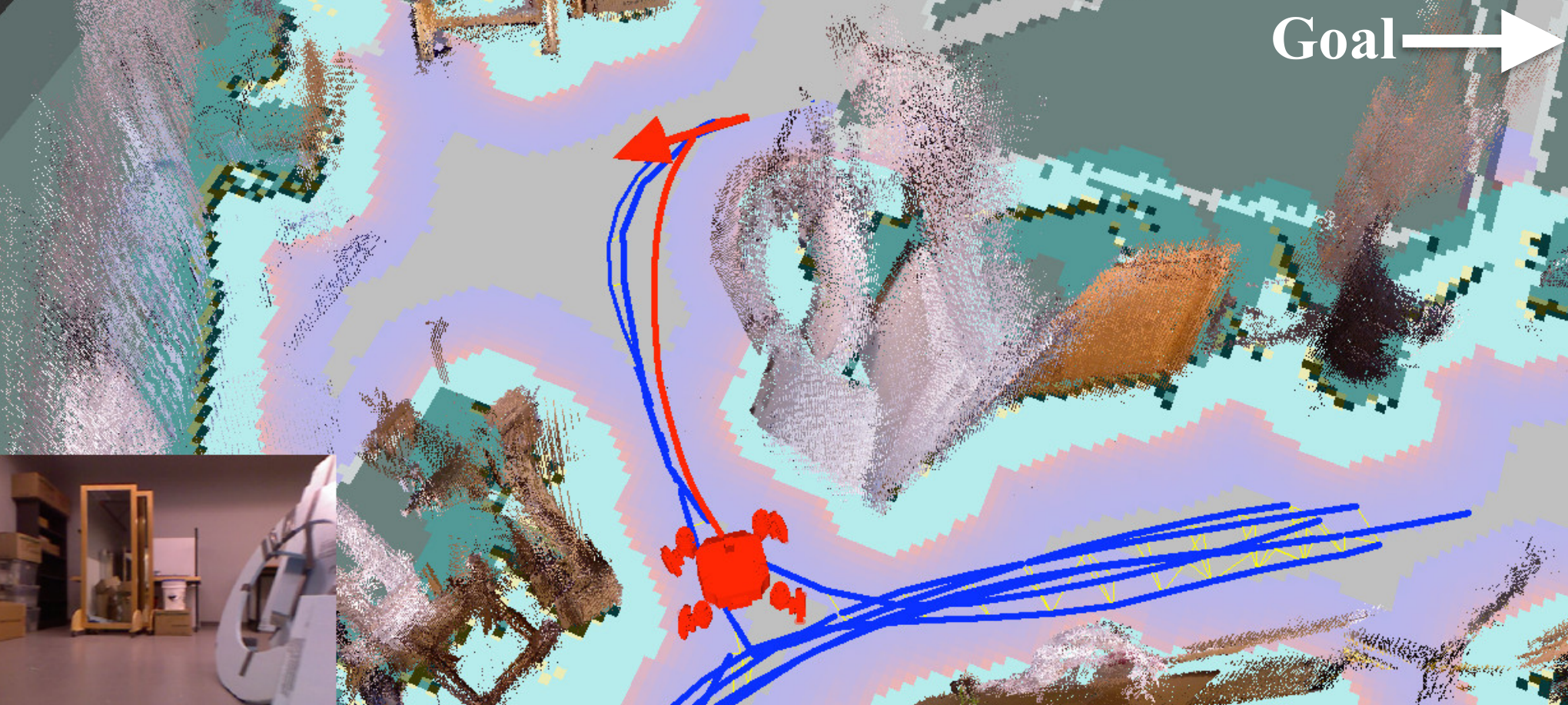}\label{fig:example_planner_b}} \\
\subfloat[]{\includegraphics[width=\columnwidth]{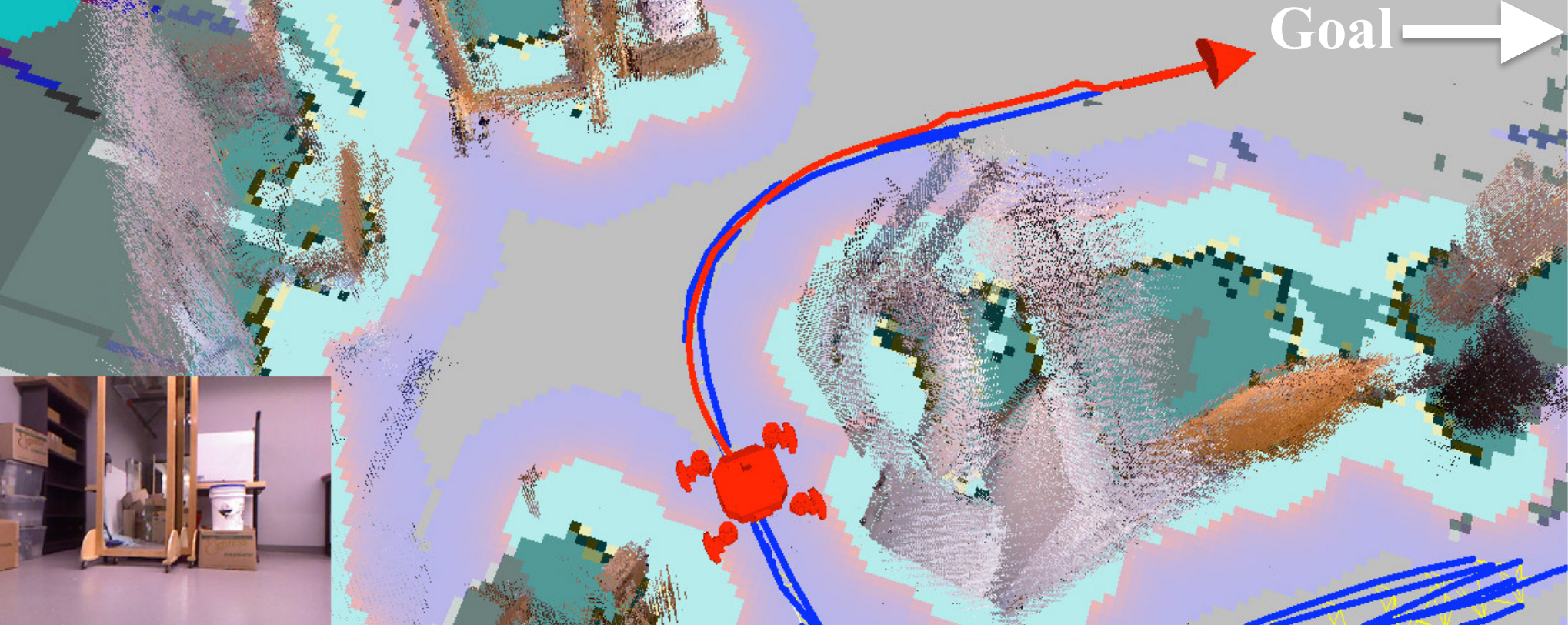}\label{fig:example_planner_c}}
\caption{Example of poses sent by TPP to MPP while nodes from LTM are retrieved for the planned path. The goal of the path is somewhere outside these images in the direction shown by Goal. The bottom left images shows the actual RGB image from the RGB-D camera. The blue lines are nodes and links of the local map. The red line is the computed trajectory from MPP using the local map's occupancy grid from its current pose (red arrow). 
The RGB point cloud and the occupancy grid are created using RGB-D images and laser scans stored in nodes from the local map, respectively.
In a), the robot is following the red trajectory. In b), some nodes are retrieved from LTM and a new trajectory is computed to move further on the path toward the goal.
} 
\label{fig:samplePlanning}
\end{figure}

To illustrate with a concrete example of the situation described in Fig. \ref{fig:example_planner},  Fig. \ref{fig:samplePlanning} presents an example of consecutive poses sent by TPP to MPP while nodes from LTM are retrieved for the planned path. 
The red arrow shows the pose of the farthest node on the path (the direction of the arrow shows the orientation of the pose). 
The red line represents the trajectory computed by MPP from the current position of the robot to its targeted pose, combined with obstacle avoidance. 
The blue lines represent the local map. 
In Fig. \ref{fig:example_planner_b}, the targeted pose is on a node traversed backward (as shown by the arrow pointing backward).
Between a) and b), the local map was updated with nodes loaded from LTM of the topological path.  
The targeted pose was updated farther on the path and at the same time, the occupancy grid was extended to previously mapped areas and MPP recomputed its trajectory. 
The robot could then move farther toward its goal and the nodes retrieved were used for proximity detection to correctly follow the planned path. 

\begin{figure}[!t] 
\centering 
\includegraphics[width= \columnwidth]{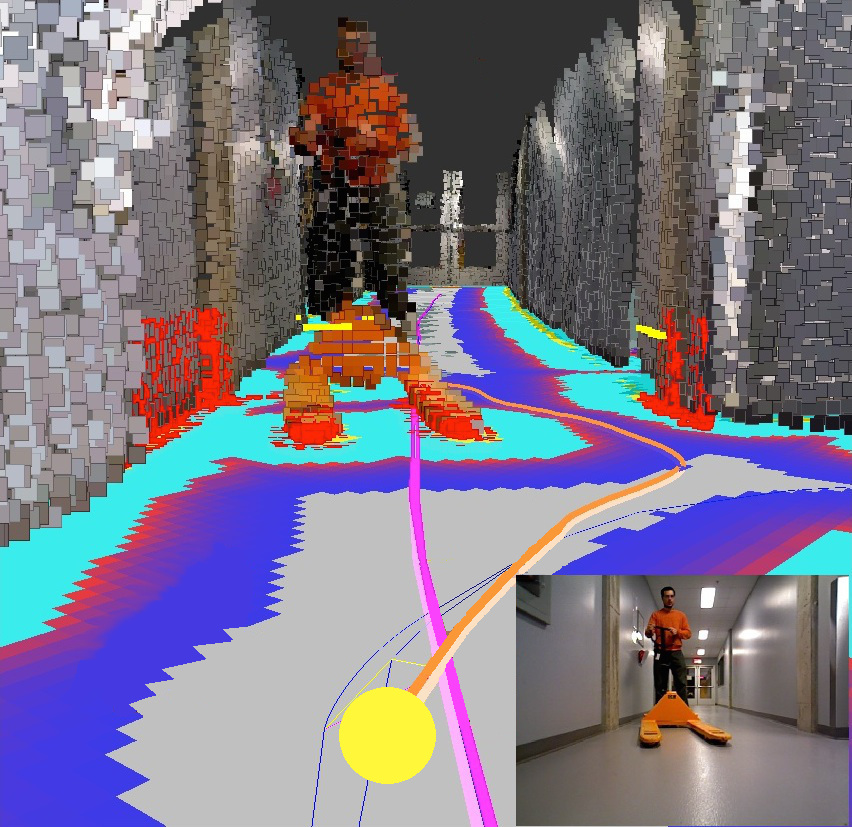} 
\caption{Example where MPP plans a slightly different path (orange) than the one provided by TPP (pink). 
The yellow dot is the current position of the robot and the lower right image is the corresponding RGB image.} 
\label{fig:avoid_obstacle} 
\end{figure}

To also illustrate the importance of obstacle detection described in Fig. \ref{fig:example_chair}, Fig. \ref{fig:avoid_obstacle} presents an example where an unexpected obstacle was encountered: as the laser rangefinder is 0.4 m above the ground, the forklift could only be detected using the RGB-D camera.
MPP planned a slightly different path (orange) that the one planned by TPP (pink) to avoid the obstacle. 

\subsection{Influences of LTM on TPP}
\label{sec:gr}
Although Fig. \ref{fig:graphAndTime} demonstrates that SPLAM-MM is able to satisfy online constraints on a map increasing linearly in size (i.e., not bounded to a maximum size of environment), memory used by LTM and consequently TPP planning time increase linearly. For example, at the end of experiment, LTM contains $24002$ nodes and $113368$ links. All raw sensor data in the nodes were also saved in the LTM's database (for debugging and visualization purposes), including RGB image (JPEG format) and depth image (PNG format) of each node. The final database took $6.7$ GB of hard drive space. With as many links at the end of the experiment, TPP required $2.4$ sec to compute a plan to the next waypoint. In term of memory usage and planning time, LTM must be somewhat limited over time when revisiting the same areas. 

\begin{figure}[!t] 
\centering 
\includegraphics[width=\columnwidth]{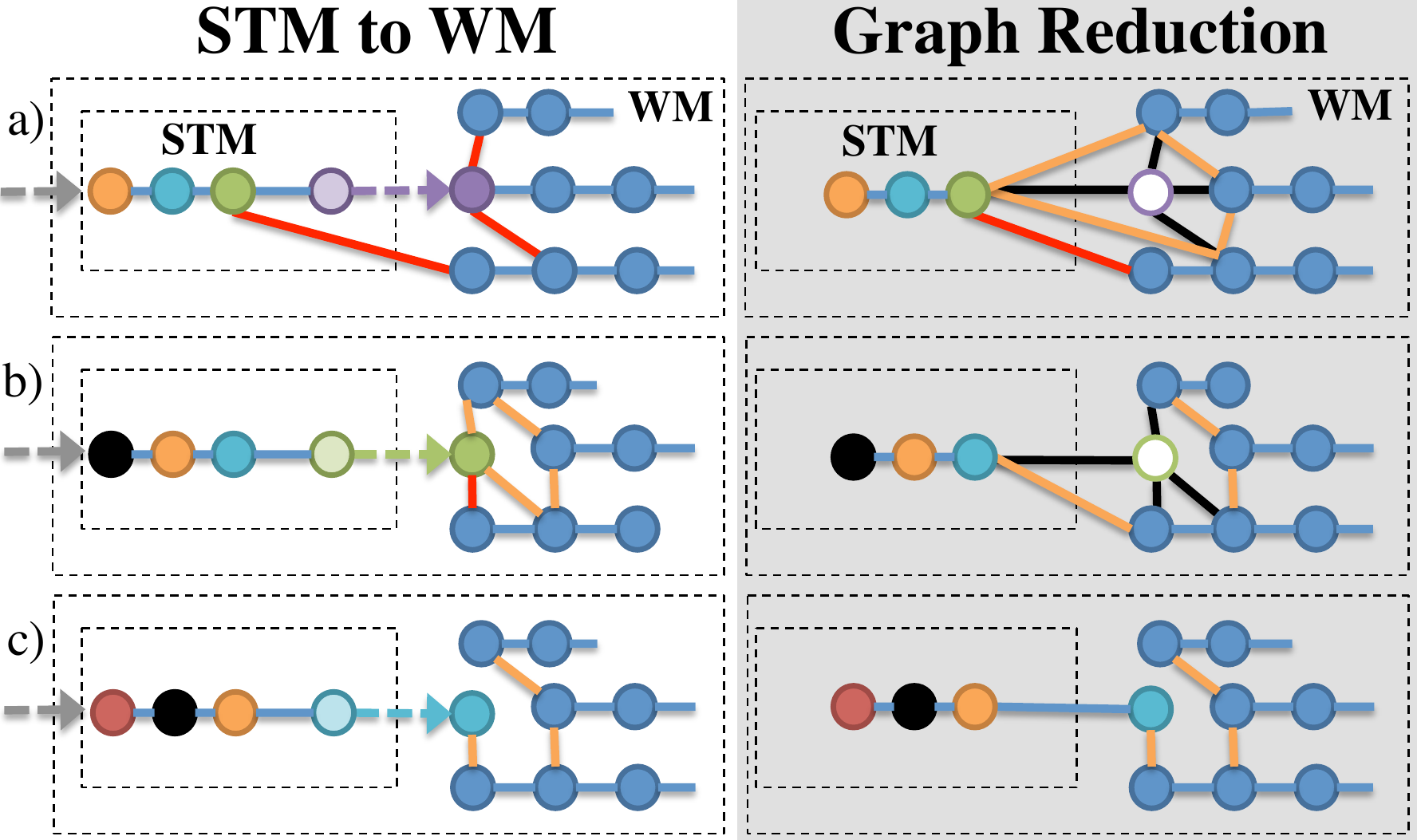} 
\caption{Three examples illustrating how the graph reduction algorithm works. Blue, red and orange links represent neighbor, loop closure and merged neighbor links, respectively. Black links and white nodes are those removed using graph reduction. The left column shows the rightmost node (the oldest) of STM moved to WM. Then on the right column, this node is removed if it has a loop closure link.} 
\label{fig:graph_reduction} 
\end{figure}

As a solution to limit LTM memory growth, nodes from STM can be merged when moved to WM if they have loop closure and/or visual proximity links. 
We studied this possibility by adding a graph reduction algorithm to STM, to remove the node from the graph and to add its neighbor links to the corresponding old node(s). 
Algorithm \ref{alg:graphreduction} summarizes the approach used to maintain the graph at the same size (same number of removed links and nodes than added) if there are many successive nodes with loop closure or visual proximity links. 
If two nodes of a same location do not have similar images (i.e., they don't have loop closure or visual proximity links), they will not be merged, thus still keeping a variety of different images representing the same location. 
To make sure nodes to be merged are still in WM (to avoid to modify the LTM), nodes having a link to a node in STM are identified as nodes that must stay in WM (similarly to Heuristic 2).
Figure \ref{fig:graph_reduction} shows how links are merged between the node moved to WM and its corresponding node(s) linked by loop closure link. 
In a), the purple node has two loop closure links. 
On graph reduction, its two neighbor links (blue) are merged with the loop closure links (red) by multiplying the corresponding transformations together, creating merged neighbor links (orange). In this case, the same number of links are added than those removed but one node is removed. 
In b), the green node has only one neighbor link (with the cyan node), then the loop closure link is only merged with it, creating only one link and four are removed. 
Merged neighbor links are ignored to be merged again to limit the number of links. 
In c), the cyan node does not have any loop closure and no graph reduction is done.

\begin{algorithm}\small
\begin{algorithmic}[1]
\STATE $o\gets$ node moved to WM
\STATE $\boldsymbol{m}\gets$ loop closure and visual proximity links of $o$
\IF{$\boldsymbol{m}$ is not empty}
\STATE $\boldsymbol{n}\gets$ neighbor links of $o$
\FORALL{$m$ in $\boldsymbol{m}$}
\STATE $o_{m}\gets$ node pointed by $m$
\FORALL{$n$ in $\boldsymbol{n}$}
\STATE $o_{n}\gets$ node pointed by $n$
\STATE $t\gets$ $m^{-1}$$\cdot$$n$ 
\STATE Add $t$ to $o_{m}$
\STATE Add $t^{-1}$ to $o_{n}$
\ENDFOR
\ENDFOR
\STATE Remove $o$ from the graph
\ENDIF
\end{algorithmic}
\caption{Graph Reduction} 
\label{alg:graphreduction} 
\end{algorithm}

To test this idea, data from the 11 sessions were processed again to test the influences of the graph reduction approach using real data acquired by the robot. 
Note that even though graph reduction was validated offline, we carefully monitored the experiment manually to make sure that the robot could still localize itself correctly on the planned paths. 

\begin{figure}[!t] 
\centering 
\includegraphics[width= \columnwidth]{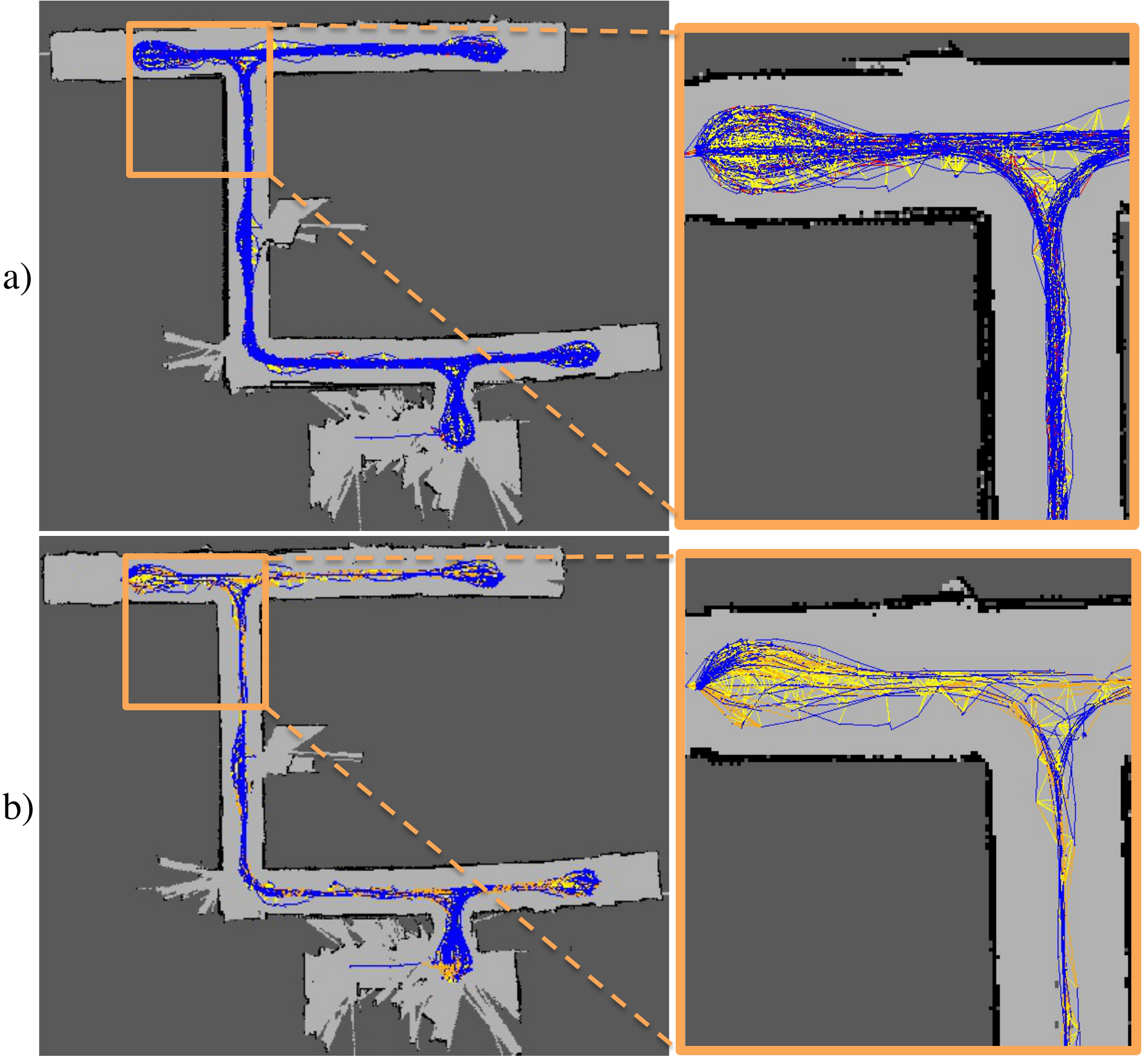} 
\caption{Comparison between the global maps a) without graph reduction ($24002$ nodes and 113368 links); b) with graph reduction (6059 nodes and 18255 links).} 
\label{fig:global_maps} 
\end{figure}

Figure \ref{fig:global_maps} shows a comparison of the final global map without and with graph reduction. 
The zones with less blue links indicate that there were many nodes merged. 
The zones with more blue links are where nodes were not merged, because of a lack of features or because of obstacles: the robot was not able to localize itself perfectly on the paths every time, thus adding new nodes to the map. 

\begin{figure}[!t] 
\centering 
\includegraphics[width= \columnwidth]{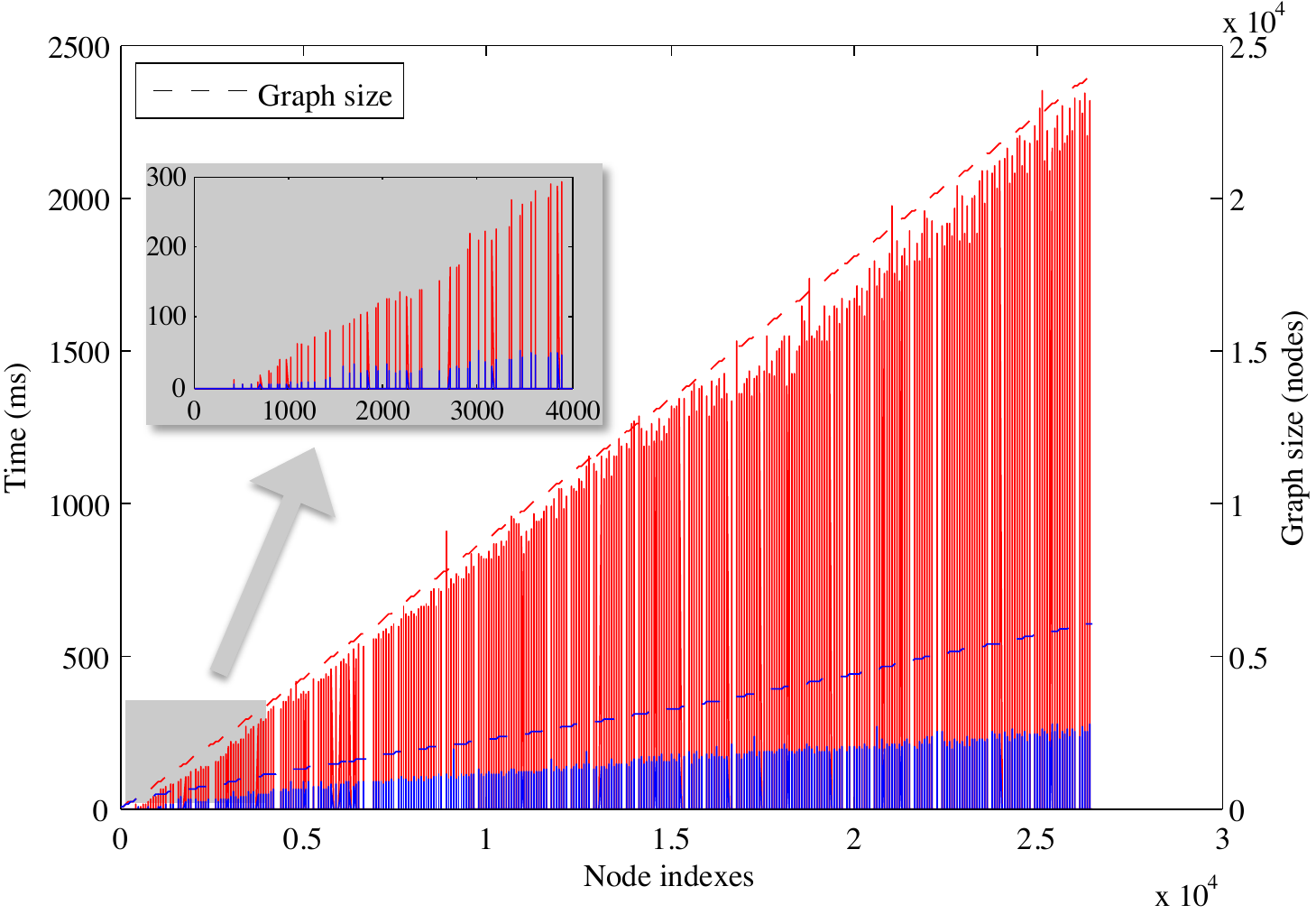} 
\caption{Comparison of TPP planning time and LTM size, with (blue) and without (red) graph reduction. The peaks in the zoomed section show more precisely when a planning is done (when a waypoint is reached).} 
\label{fig:globalGraphPlanningTime} 
\end{figure}

Figure \ref{fig:globalGraphPlanningTime} illustrates TPP planning time corresponding to LTM size with and without graph reduction. 
As the LTM became larger, TPP planning time increased: with graph reduction, TPP planning time was reduced by 89\% for the last path planned ($272$ ms instead of $2.4$ sec). 
Figure \ref{fig:globalGraphMB} illustrates hard drive usage with and without graph reduction. 
Extrapolating linearly memory usage with a 100 Gb hard drive, the robot could navigate online approximately 110 hours without graph reduction before filling up the hard drive. When debugging data (not used for navigation) are not recorded in the database, this estimate would increase to approximately 33 days (800 hours). This means that if the robot is always visiting new locations at a mean velocity of 1.4 km/h (as in this experiment), it could travel up to 1120 km to map environments online. When graph reduction is used, debugging data are not saved and having the robot always revisiting the same areas like in this experiment, it could do SPLAM continuously for about 130 days before reaching the hard drive capacity.

\begin{figure}[!ht]
\centering
\includegraphics[width=\columnwidth]{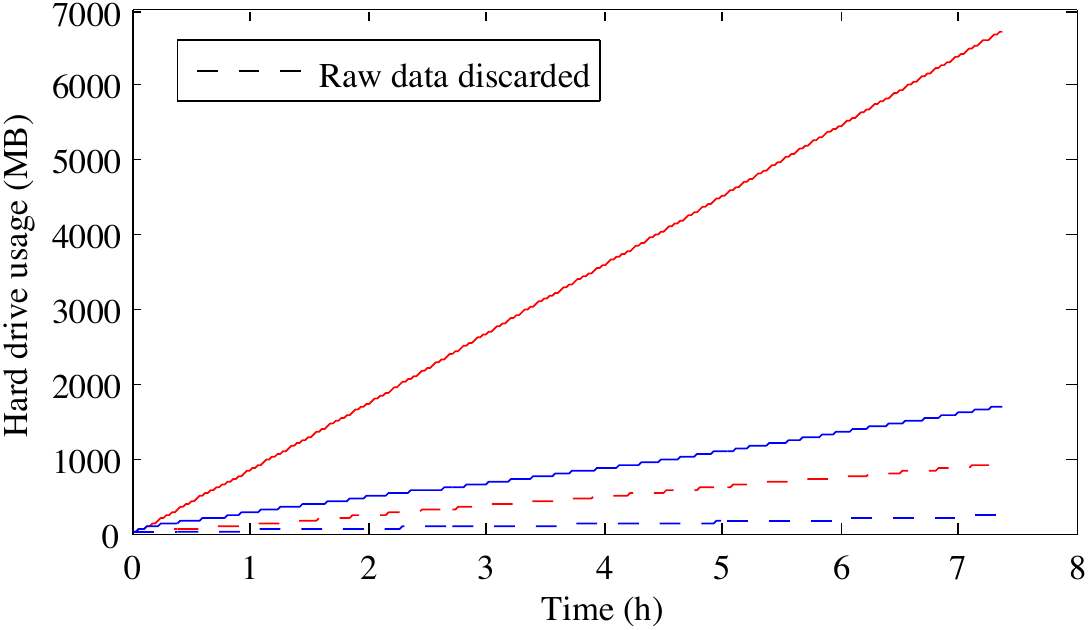}\label{fig:globalGraphMB}
\caption{Comparison of hard drive usage with (blue) and without (red) graph reduction. The dashed curves represents results without saving in database the debugging data (i.e., raw RGB and depth images). }
\label{fig:globalGraphMB}
\end{figure}

\section{Discussion}
\label{sec:discussion}
In terms of processing time, results show that SPLAM-MM is able to satisfy online processing requirements independently of the size of the environment, by transferring in LTM portions of the map which then cannot be used for loop closure detection, proximity detection and graph optimization. 
Results show also that path following is still possible in such conditions by incrementally retrieving locations on the planned path. Thus, as shown in Section \ref{sec:gr}, the current hardware limitation of the system for long-term continuous SPLAM is hard drive capacity, not computation power.

To successfully follow a path, results demonstrate the importance of adding loop closure and/or proximity links with nodes on the planned path to localize the robot in the map. 
In our trials, the robot navigated indoor where static structures (e.g., walls) were most of the time visible using the laser rangefinder. 
However, in large empty spaces where the laser rangefinder would not be able to perceive nearby structures, it would be difficult for the robot to follow a path if appearance-based loop closure detection and visual proximity detection do not occur. 
A laser rangefinder with larger perceptual range or a 3D LIDAR sensor like the Velodyne could be used to increase perceptual range.
For a lower cost solution, using a camera facing backward could be useful to allow the robot to detect similarities in images when traversing a path in opposite direction \citep{carrera2011lightweight}. 
Without adding new sensors, TPP could also stop sending new poses when no loop closure links or proximity links occur for a while. 
If no loop closures were found over the next few meters, it would be possible to wait for the robot to rotate at this location so that it can look backward, increasing its chance to detect a loop closure to correct its position on the planned path and then generate a new pose. 
A similar recovery approach is presented in \citep{milford2010persistent}, where an exploration phase is triggered to re-localize the robot when failing to follow the planned path. 
Also, to be more robust to dynamic environments where there are cyclic changes over time, TPP could select nodes that match better the current time of the day rather than the most recent ones, to increase localization success as in \citep{krajnik2016persistent}.

In comparison with large empty environments, those in which a lot of dynamic changes occur (e.g., navigating through a crowd) would also make simultaneous planning and localization more difficult. 
For instance, mapping the area in session 1 without people walking by helped the robot acquire the static structures of the environment since they were not hidden by people. 
These static structures facilitate localization when the robot comes back to these areas later one. 
If these static structures were previously occluded, they would be added to the map as the robot comes back to these areas (obviously if people are no longer in the robot's field of view).
If people partially occlude the robot's sensors over a long distance, localization would still be possible but would occur less frequently. 

For online multi-session mapping with our memory management approach, the worst case is when all nodes of a previous map are transferred to LTM before a loop closure is detected \citep{labbe13appearance}. 
This results in definitely ignoring the previous map and disabling at the same time the ability to plan paths to a location in it. 
To avoid this problem, an additional heuristic could be to keep in WM at least one discriminative node for each map. 
However, if the number of mapping sessions becomes very high (e.g., thousands of sessions), these nodes would definitely have to be transferred in LTM to satisfy online processing requirements. 
A strategy that makes the robot explore potential paths to link maps together would then be useful, and maps that could not be linked would eventually be unretrievable. 

In the trials conducted, no invalid loop closures were detected, avoiding to corrupt the map with erroneous loop closure links. 
If this happens, graph optimization approaches such as \citep{latif2013robust, sunderhauf2012towards, lee13robust} deal with possible invalid matches, and could be used to increase robustness of SPLAM-MM. 
However, these approaches assume that the whole global map is available online, which is not the case here. 
They could be still used offline at the end of a session.

As shown by Fig. \ref{fig:avoid_obstacle}, MPP in SPLAM-MM allows the robot to find an alternative path to reach the targeted pose when possible. 
However, if the alternative path is outside the local map, re-planning with TPP is required.
Some paths may be also blocked temporary or permanently by some dynamic or new static obstacles. 
An approach similar to \citep{konolige2011navigation} could be used to identify some links as blocked so that TPP cannot plan a path using them. 
The Patrol module could also manage waypoints that can and cannot be reached.

Finally, the graph reduction approach can reduce significantly the number of nodes and links saved in LTM to reduce TPP planning time. However, because of dynamic events or the lack of features (e.g., Fig. \ref{fig:obstacles}e), new nodes and links will inevitably be added to LTM over time when revisiting the same areas. 
As an improvement, nodes with featureless image could be merged through a maximum density threshold like in \citep{milford2010persistent}, as they cannot be used for loop closure detection. 
After applying graph reduction on the experimental data, there are still 3068 featureless nodes of 6059 nodes in the global graph, which would reduce by about 50\% the remaining graph. 
However, even by limiting the rate at which the LTM grows, a continuous SLAM approach in unbounded dynamic environments will always add new data over time. 
A complementary strategy would be to definitely forget some parts of the global map, at the cost of not being able to return to some locations.

\section{Conclusion}
\label{sec:conclusion}

By limiting the nodes of the map available online in WM for loop closure detection, proximity detection and graph optimization, results presented in this paper suggest that the proposed graph-based SPLAM-MM approach is able to meet online processing requirements needed for simultaneous mapping, localizing and planning in multi-session conditions.  
SPLAM-MM is tightly based on appearance-based loop closure detection, allowing it to naturally deal with the initial state problem of multi-session mapping. 
To successfully localize on a planned path through areas previously transferred in LTM, memory management allows SPLAM-MM to deal with the necessity of retrieving upcoming nodes on the path in WM.
Our code is open source and available at \url{http://introlab.github.io/rtabmap}. 

In future works, more robust failure recovery approaches will be examined to test SPLAM-MM in dynamic environments where the paths could often be blocked (temporally or permanently). We also plan to study the impact of autonomous coverage and exploration strategies, especially how it can actively direct exploration based on nodes available for online mapping. This could be also useful to conduct longer experiments at larger scale.


\bibliographystyle{spbasic}      
\bibliography{main.bib}   

\end{document}